\documentclass[11pt]{article}
\usepackage{eccv24}
\usepackage[utf8]{inputenc} 
\usepackage[T1]{fontenc}    
\usepackage{hyperref}       
\usepackage{url}            
\usepackage{booktabs}       
\usepackage{amsfonts}       
\usepackage{nicefrac}       
\usepackage{microtype}      
\usepackage{lipsum}
\usepackage{amsmath}
\usepackage{algorithm}
\usepackage{algpseudocode}
\usepackage{graphicx}       
\graphicspath{{media/}}     

\usepackage[utf8]{inputenc}
\usepackage{authblk}         
\usepackage{geometry}        
\usepackage{graphicx}
\usepackage{caption}
\usepackage{subcaption}
\usepackage{amsmath, amssymb}
\usepackage{hyperref}
\usepackage{xcolor}
\usepackage{booktabs}        
\usepackage[normalem]{ulem}  

\usepackage{booktabs}  
\usepackage{graphicx}
\usepackage{caption}
\usepackage{subcaption}
\AtBeginDocument{%
  \providecommand\BibTeX{{%
    \normalfont B\kern-0.5em{\scshape i\kern-0.25em b}\kern-0.8em\TeX}}}

\usepackage{graphicx}
\usepackage{caption}
\usepackage{subcaption}

\usepackage[normalem]{ulem}
\usepackage{xcolor}
\usepackage{pifont}
\newcommand{\cmark}{\ding{51}}%
\newcommand{\xmark}{\ding{55}}%

\usepackage{ragged2e}

\begin{document}




\title{DANCER: Dance Animation via Condition Enhancement and Rendering with Diffusion Model}

\author{Yucheng Xing$^{*}$, Jinxing Yin$^{*}$, Xiaodong Liu$^{*}$, Xin Wang\\
\small Stony Brook University, New York, USA\\
\small \texttt{\{yucheng.xing, jinxing.yin, xiaodong.liu, x.wang\}@stonybrook.edu}\\

\vspace{0.5em}
\small $^{*}$Equal contribution.}
\date{}

\maketitle


\begin{abstract}


Recently, diffusion models have shown their impressive ability in visual generation tasks. Besides static images, more and more research attentions have been drawn to the generation of realistic videos. The video generation not only has a higher requirement for the quality, but also brings a challenge in ensuring the video continuity. Among all the video generation tasks, human-involved contents, such as human dancing, are even more difficult to generate due to the high degrees of freedom associated with human motions. In this paper, we propose a novel framework, named as \textit{\textbf{DANCER} (\textbf{D}ance \textbf{AN}imation via \textbf{C}ondition \textbf{E}nhancement and \textbf{R}endering with Diffusion Model)}, for realistic single-person dance synthesis based on the most recent stable video diffusion model. As the video generation is generally guided by a reference image and a video sequence,
we introduce two important modules into our framework to fully benefit from the two inputs. More specifically, we design an \textit{\textbf{Appearance Enhancement Module (AEM)}} to focus more on the details of the reference image during the generation, and extend the motion guidance through a \textit{\textbf{Pose Rendering Module (PRM)}} to capture pose conditions from extra domains. To further improve the generation capability of our model, we also collect a large amount of video data from Internet, and generate a novel dataset \textit{\textbf{TikTok-3K}} to enhance the model training. The effectiveness of the proposed model has been evaluated through extensive experiments on real-world datasets, where the performance of our model is superior to that of the 
state-of-the-art methods. All the data and 
codes will be released upon acceptance. 

\end{abstract}
\begin{center}
\end{center}


\section{Introduction ~\label{sec:1}}

Creativity is one of the most important factors to measure the machine intelligence. In recent years, the emergence of various generative models, such as Variantional Auto-Encoders (VAEs)~\cite{pinheiro2021variational}, Generative Adversarial Networks (GANs)~\cite{goodfellow2020generative} and Diffusion Models (DMs)~\cite{ho2020denoising, song2020denoising}, has inspired more and more researchers to focus on the generation tasks, 
and the quality of the generative models reflect the creativity level of the machine learning algorithms. From images to videos, the objects to be generated 
get increasingly complex, benefiting from the complicated model structures and powerful computational resources. Among the generated contents, human-related activities, such as human dances, are particularly difficult due to the high degrees of freedom of human motions
while they are also of vital value for building better human-assisted models in real applications. 


Currently, there are several challenges that 
limit the generation performance of human dance videos. On one hand, due to the 
variety of changes in human body movements and the insufficient information provided in the reference image, 
details such as occluded limb distal segments and surrounding backgrounds 
may have been lost or distorted when transferred from the reference image to the corresponding video frames. 
On the other hand, besides the generation quality of a single video frame, the continuity of the adjacent frames is also of vital importance. Since the video frames are generated under the guidance of the prior pose sequence, the limited information and inherent jitters of the pose guidance will 
impact the smoothness of the generated videos. 

In the human dance synthesis field, most current works rely on the generative models, such as GANs~\cite{sarkar2021humangan, siarohin2019first, tian2021good, wang2020g3an, yoon2021pose} and DMs~\cite{ma2024follow, feng2023dreamoving, karras2023dreampose, wang2024disco, hu2024animate, chang2023magicpose, wang2024vividpose, zhang2024mimicmotion, zhai2024idol}. Compared to GANs, DMs obtain a relatively better performance in general with a more stable training process. 
Existing DM-based methods 
usually integrate the reference image and the pose sequence into the diffusion process as extra conditions to make sure that the generated dance videos can match them. 
However, most of these methods overlook the subtle details present in reference images during generation and are limited by the lack of rich information in skeleton maps used for pose guidance. 


To address the limitations in current models for human dance synthesis, we propose a novel framework named \textit{\textbf{DANCER}} for high-quality generation of single-person dance videos. Our approach introduces two key modules:  1) a detail-aware \textit{Appearance Enhancement Module (AEM)}, which captures subtle features from the reference image to enhance the visual quality of each generated frame; and 2) a \textit{Pose Rendering Module (PRM)}, which enhances pose guidance to provide more detailed and accurate motion cues by enriching sparse skeleton maps with additional information extracted from the source reference video, such as segmentation maps, depth maps, and 3D parametric models. Together, AEM and PRM allow DANCER to generate dance videos with improved realism, visual fidelity, and temporal coherence. 
Another critical challenge in single-person dance generation is the limited size and diversity of existing datasets, which hampers the training effectiveness of deep learning models. To address this, we collect a wide variety of dance videos from the Internet and create a new large-scale dataset, \textbf{TikTok-3K}. This dataset significantly expands the available training data and is designed to support more robust and generalizable learning, which will ultimately benefit future research and development in the field of dance video generation. Our contributions can be summarized in the following four aspects:
\begin{itemize}
    \item We propose a new SVD-based framework for the generation of single-person dance video, 
    and verify the effectiveness of the framework 
    through extensive experiments.
    \item We design a novel appearance enhancement module 
    to focus more on the details during generation, which enhances the extracted appearance features 
    on different levels. 
    \item In order to improve the smoothness of the generated video, we augment the motion guidance and provide more pose references instead of only using skeleton maps, where 
    motion conditions 
    are preprocessed to remove jitters before 
    being used in the generation to maintain the smoothness.  
    \item 
    We construct a new dataset comprising a large collection of human dance videos sourced and processed from the Internet. This dataset significantly extends the volume and diversity of existing resources in the field, providing a richer foundation for training and evaluating high-quality dance video generation models.
\end{itemize}

The remainder of this paper is organized as follows. In Sec.~\ref{sec:2}, we provide a brief review of the previous work related to human dance synthesis and diffusion models. In Sec.~\ref{sec:6} and Sec.~\ref{sec:3}, we introduce the large dataset we newly create to facilitate the training of sophisticated network, and  present in details our model  designed for generating high quality video. We evaluate the effectiveness of the proposed framework 
through extensive experiments in Sec.~\ref{sec:4}. Finally, we conclude our work and  point out some future research directions in Sec.~\ref{sec:5}.



\section{Related Works~\label{sec:2}}


Human dance synthesis, or human image animation, targets at creating a consecutive video frames containing dance-related contents based on a static image and a sequence of poses. Previously, GAN-based methods~\cite{sarkar2021humangan, siarohin2019first, tian2021good, wang2020g3an, yoon2021pose} mainly employ spatial warping functions to transform the reference image according to the expected video frames 
along with the predicted flow map extracted from the motion changes of the pose sequence. However, the performance of this branch of methods 
is often compromised with insufficient information caused by mode collapse. 
This limits their capability in dealing with local details, keeping temporal consistencies and recovering the missing parts from occlusion. Recently, with the emergence of Diffusion Model (DM)~\cite{croitoru2023diffusion}, a much more powerful generative model superior to GANs on generation quality and training stability, DM-based methods~\cite{ma2024follow, feng2023dreamoving, karras2023dreampose, wang2024disco, hu2024animate, chang2023magicpose, wang2024vividpose, zhang2024mimicmotion, zhai2024idol} have also been proposed to generate human dance videos 
under the supervision of a given pose sequence. Specifically, the reference image and the pose sequence 
are used as two conditions during the frame generation process. In DreaMoving~\cite{feng2023dreamoving}, the reference image is directly encoded by CLIP~\cite{radford2021learning} to replace the original text prompts to guide the Stable Diffusion~\cite{rombach2022highresolution}. However, since CLIP is trained to match high-level semantic features for text, 
it lacks the capability of capturing low-level details 
and is thus not enough to get a good reference condition. To solve this problem, DreamPose~\cite{karras2023dreampose} combines CLIP with VAE in their encoder, while DisCo~\cite{wang2024disco} processes the foreground and the background of the reference image separately with the help of ControlNet~\cite{zhang2023adding}. AnimateAnyone~\cite{hu2024animate} further extends this idea and designs a novel ReferenceNet, which is also used in MagicPose~\cite{chang2023magicpose}. Besides, VividPose~\cite{wang2024vividpose} observes the limitation of identity consistency in generated video and adds additional id controller with a reference face. As for the pose guidance, a skeleton sequence drawn from pose estimators, such as OpenPose~\cite{cao2017realtime} or DWPose~\cite{yang2023effective}, is commonly used in~\cite{wang2024disco, hu2024animate, chang2023magicpose}. However, the extracted skeletons might be inaccurate, and MimicMotion~\cite{zhang2024mimicmotion} proposes to use detection confidence to enhance the extracted skeleton map in the condition embedding. To ease the sparsity problem in these skeletons, MagicAnimate~\cite{xu2024magicanimate} adopts DensePose~\cite{guler2018densepose} while some works~\cite{wang2024vividpose, zhu2024champ} even utilize 3D parametric SMPL model~\cite{loper2023smpl}.



\section{TikTok-3K Dataset~\label{sec:6}}

\begin{figure*}[!htpb]
    \centering
    \includegraphics[width=\linewidth]{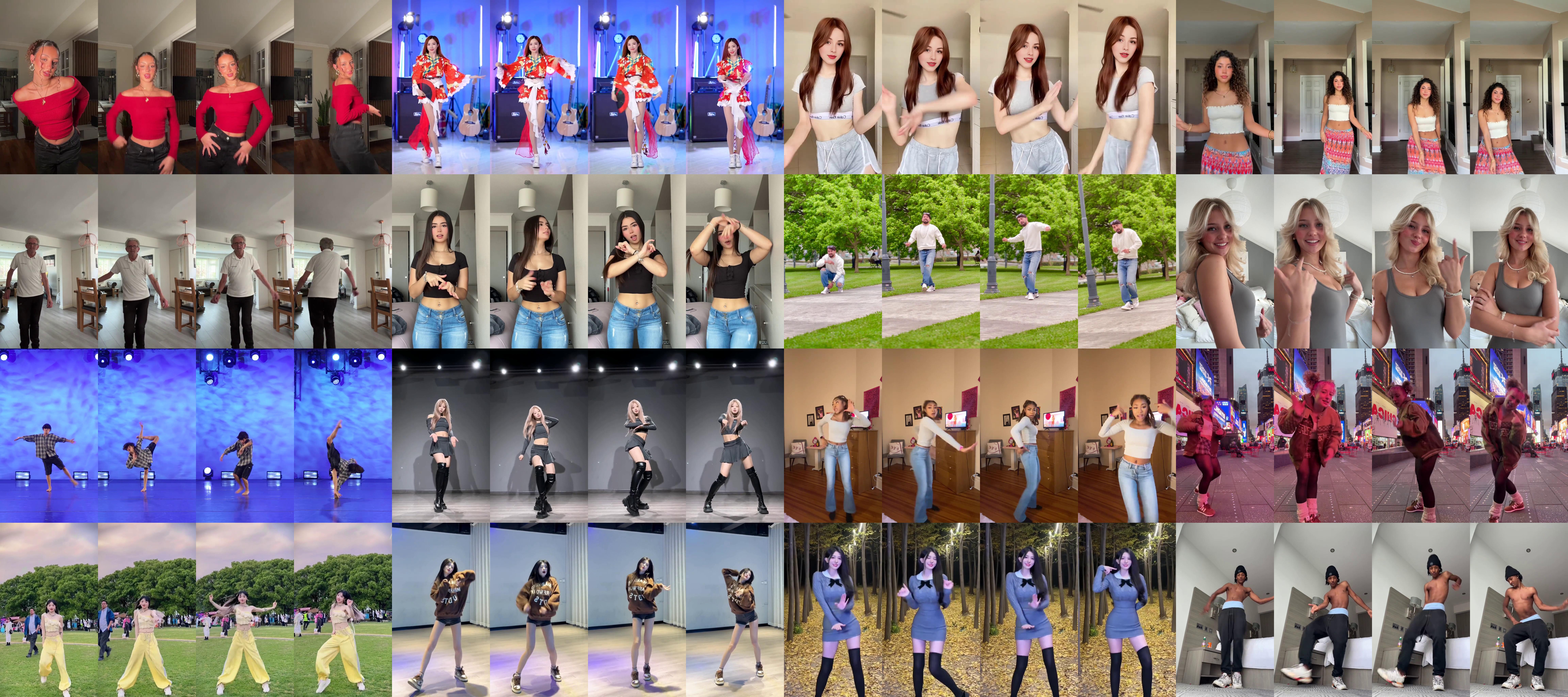}
    \caption{Illustration of video data in our \textit{TikTok-3K} dataset.}
    \label{fig:dataset_1}
\end{figure*}

Before diving into the details of our proposed model, we 
would like to first introduce our newly created large-scale single-dancer video dataset, \textit{TikTok-3K}.


Modern machine learning models, particularly complex generative models with extensive learnable parameters, demand substantial amounts of high-quality training data to achieve a satisfactory performance. 
However, for the single-person dance generation, there exists only one commonly used dataset, $TikTok$~\cite{jafarian2021learning}. It contains about $350$ single-person dance videos, a very small number that limits its capacity for the training of diffusion models for video generation. The videos in $TikTok$ last for $10$-$15$ seconds, with the sampling rate of $30$ frames per second. Most videos record the faces and upper-bodies of dancers, and some have very low resolutions. 
The lack of large, high-resolution datasets with diverse video samples motivates us to curate TikTok-3K. Our dataset comprises approximately 3,000 carefully selected video clips with varied lengths, primarily ranging from 10 to 20 seconds, alongside some longer clips for learning long-range dependencies. These longer clips, typically recording the performance of professional dancers, feature complex and intricate motions.

TikTok-3K also offers greater variety across dance types, including genres from casual to classical, and expands on the environmental diversity. While prior datasets primarily consist of videos recorded in static indoor settings, TikTok-3K includes clips filmed in both static and dynamic outdoor environments, supporting model adaptability to different conditions. Additionally, our dataset encompasses a broad representation of dancers across races, genders, and age groups (Fig.~\ref{fig:dataset_1}), which helps prevent model overfitting to specific distributions—a limitation observed in previous generative models.

\begin{figure}[t]
    \centering
    \includegraphics[width=\linewidth]{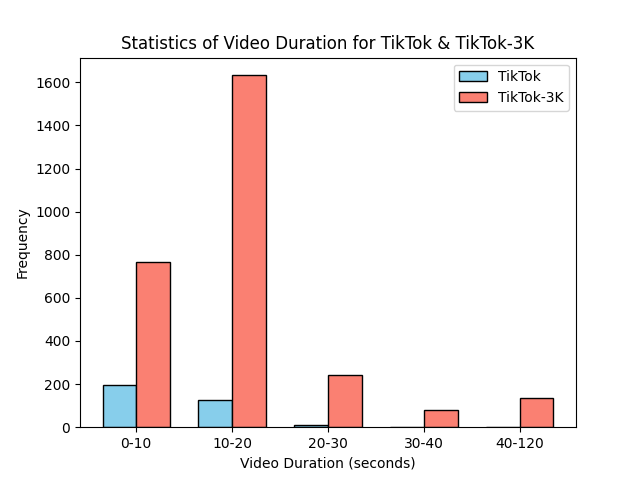}
    \caption{Comparison of frame distribution between \textit{TikTok-3K} and \textit{TikTok}.~\label{fig:dataset_comp}}
\end{figure}

To ensure the usability, all videos have been 
manually selected and preprocessed. This process is critical and introduced to make sure that each clip is relevant, without blank or extraneous content, and ready for training. For better comparison with prior datasets, we present a breakdown of frame distributions between TikTok-3K and the traditional TikTok dataset, visualized in Fig.~\ref{fig:dataset_comp}.


\section{Methodology~\label{sec:3}}

\begin{figure*}[!htpb]
    \centering
    \includegraphics[width=0.95\linewidth]{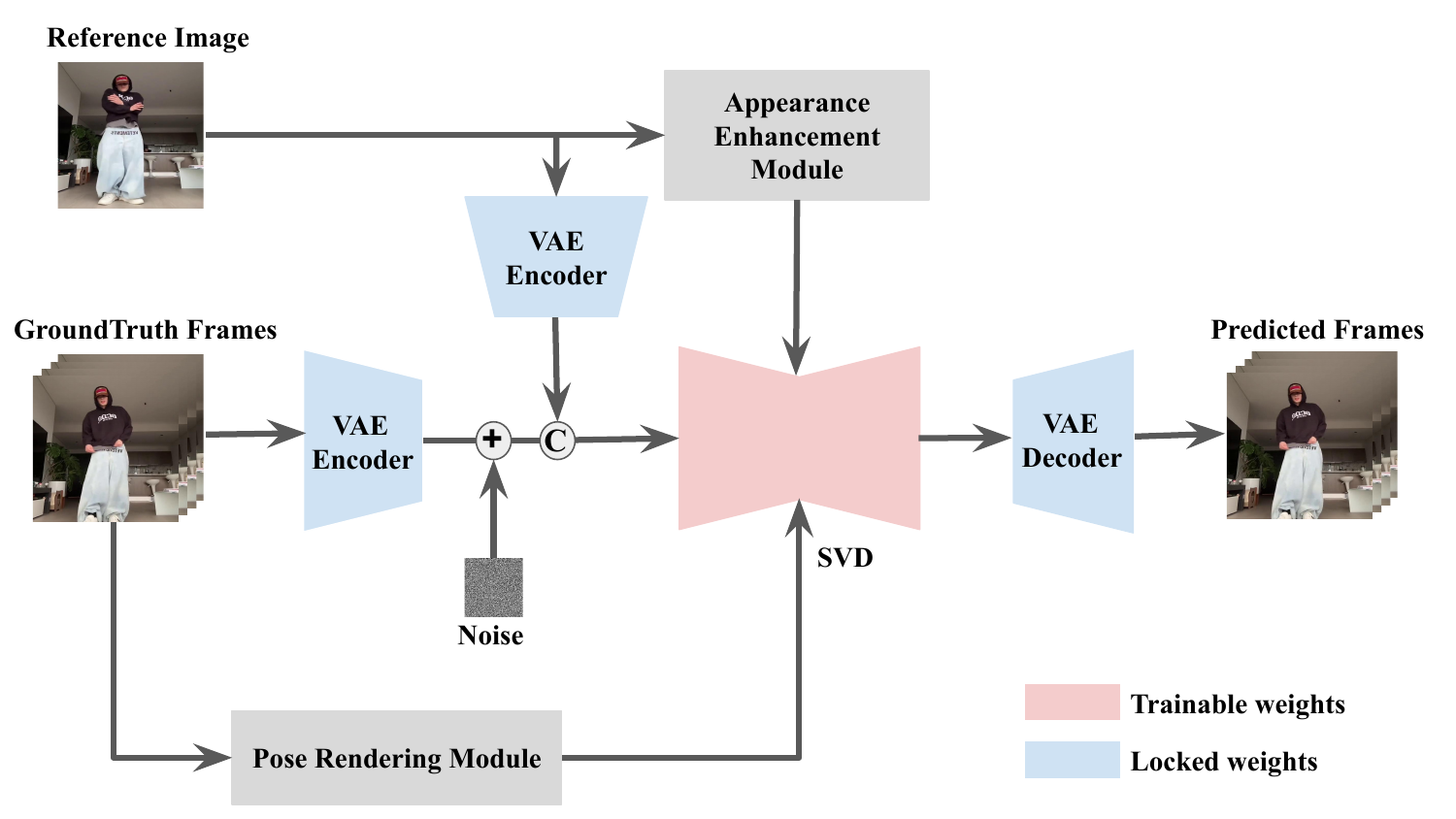}
    \caption{Overall framework of our DANCER model.~\label{fig:framework}}
\end{figure*}

In this section, we will first give an overview of the framework of our \textit{DANCER} model, and then introduce the 
two new modules, \textit{Appearance Enhancement Module (AEM)} and \textit{Pose Rendering Module (PRM)}, separately in detail.  

\subsection{Notations \& Objective~\label{sec:3.1}}

Given an $N$-frame dance video $V_{src} = \{I_{src, 0}, I_{src, 1}, ...,$ $I_{src, N}\}$ of the source performer $src$, where $I_{src, i}$ is the $i^{th}$ frame within it, and a reference image $I_{tgt}$ of the target person $tgt$, our goal is to design an effective model with the function $f(\cdot)$ to generate the corresponding $N$-frame video $V_{tgt} = \{I_{tgt, 0}, I_{tgt, 1}, ...,$ $I_{tgt, N}\}$ whose subject is $tgt$, i.e.
\begin{equation}
    f(V_{src}, I_{tgt}) := V_{tgt}.~\label{eq:obj}
\end{equation}


\subsection{Framework~\label{sec:3.2}}

As shown in Fig.~\ref{fig:framework}, our model adopts a Latent-Diffusion~\cite{rombach2022high} framework, where video frames are first mapped into a low-dimensional latent space 
through a pre-trained VAE~\cite{pinheiro2021variational}.  
The diffusion process is 
then carried out within this latent space to generate the required feature maps. 
Unlike image generation, video synthesis 
poses significantly higher computational demand due to the high-dimensional property of the video data. 
By operating in a compressed latent space, our approach effectively reduces the computational cost while maintaining the fidelity of the generated content.

Specifically, given an image data $I_{in}$, the encoder $\mathbf{E}_{vae}(\cdot)$ 
within the VAE will extract an abstract feature $K$, which can be perfectly restored to a pixel-level image $I_{out}$ through the VAE decoder $\mathbf{D}_{vae}(\cdot)$ accordingly, with $I_{in} = I_{out}$, i.e. 
\begin{equation}
    I_{in} = I_{out} = \mathbf{D}_{vae}(K) = \mathbf{D}_{vae}(\mathbf{E}_{vae}(I_{in})).~\label{eq:vae}
\end{equation}
Therefore, to obtain the $i^{th}$ frame $I_{tgt, i}$ of the expected dance video, we only need to make sure that the corresponding latent feature $K_{tgt, i}$
can be well generated through an conditional diffusion model $G(\cdot)$. 
The diffusion process is guided by both conditions $c_{app}$ and $c_{pose}$, 
which are provided by the target reference image $I_{tgt}$ and the source dance video $V_{src}$ through our newly designed \textit{Appearance Enhancement Module (AEM, $\mathbf{F}_{AEM}(\cdot)$)} and \textit{Pose Rendering Module (PRM, $\mathbf{F}_{PRM}(\cdot)$)} separately. 
The details of the two modules will be discussed later in this section ~\ref{sec:3.3} and ~\ref{sec:3.4}.  

For the diffusion model $\mathbf{G}(\cdot)$, we choose to use the Stable Video Diffusion (SVD)~\cite{blattmann2023stable}, a powerful image-to-video diffusion model trained on a large-scale video dataset. Compared to previous video generation models, SVD has shown superior performance on both generation quality and continuity, by adding additional temporal cross-attention layers right after the spatial convolutional layers within the U-Net~\cite{ronneberger2015u} to handle the temporal consistency among generated latent features of adjacent frames. Following the formulations in~\cite{blattmann2023stable}, the generation process of the target latent features $S_{tgt, t} = \{K_{tgt, 0, t}, K_{tgt, 1, t}, ..., K_{tgt, N, t}\}$ in the $(T - t)^{th}$ diffusion step can be expressed as
\begin{equation}
    \begin{aligned}
        S_{tgt, t} &= \mathbf{G}(S_{tgt, t + 1}, c_{app}, c_{pose}, t), \\ 
        c_{app} &= \mathbf{F}_{AEM}(I_{tgt}), \\ 
        c_{pose} &= \mathbf{F}_{PRM}(V_{src}),~\label{eq:diffusion} 
    \end{aligned}
\end{equation}
where $T$ is the total number of 
diffusion steps. To better integrate the information of 
the target reference image $I_{tgt}$, instead of setting the initial input $S_{tgt, T}$ of the SVD as pure Gaussian noise, we repeat the latent feature $K_{tgt}$ of the reference image 
$N$ times and concatenate them with the random noise. 
Gaussian noise is added on top of 
the repeated $K_{tgt}$ to make sure that the generated frames are similar to but not totally the same as the reference image, so
\begin{equation}
    \begin{aligned}
        &S_{tgt, T} &= \{&K_{tgt, 0, T}, K_{tgt, 1, T}, ..., K_{tgt, N, T}\}, \\
        &&= \{&Concat[\epsilon_{src, 0}, K_{tgt} + \epsilon_{tgt, 0}], \\
        &&&Concat[\epsilon_{src, 1}, K_{tgt} + \epsilon_{tgt, 1}], \\
        &&&..., \\
        &&&Concat[\epsilon_{src, N}, K_{tgt} + \epsilon_{tgt, N}]\},~\label{eq:initial_diffusion}
    \end{aligned}
\end{equation}
where $K_{tgt} = \mathbf{E}_{vae}(I_{tgt})$, 
$\{\epsilon_{src, 0}, \epsilon_{src, 1}, ...,$ $\epsilon_{src, N}\}$ and $\{\epsilon_{tgt, 0}, \epsilon_{tgt, 1}, ...,$ $\epsilon_{tgt, N}\}$ are Gaussian noise.

\subsection{Appearance Enhancement Module (AEM)~\label{sec:3.3}}

\begin{figure}[t]
    \centering
    \includegraphics[width=\linewidth]{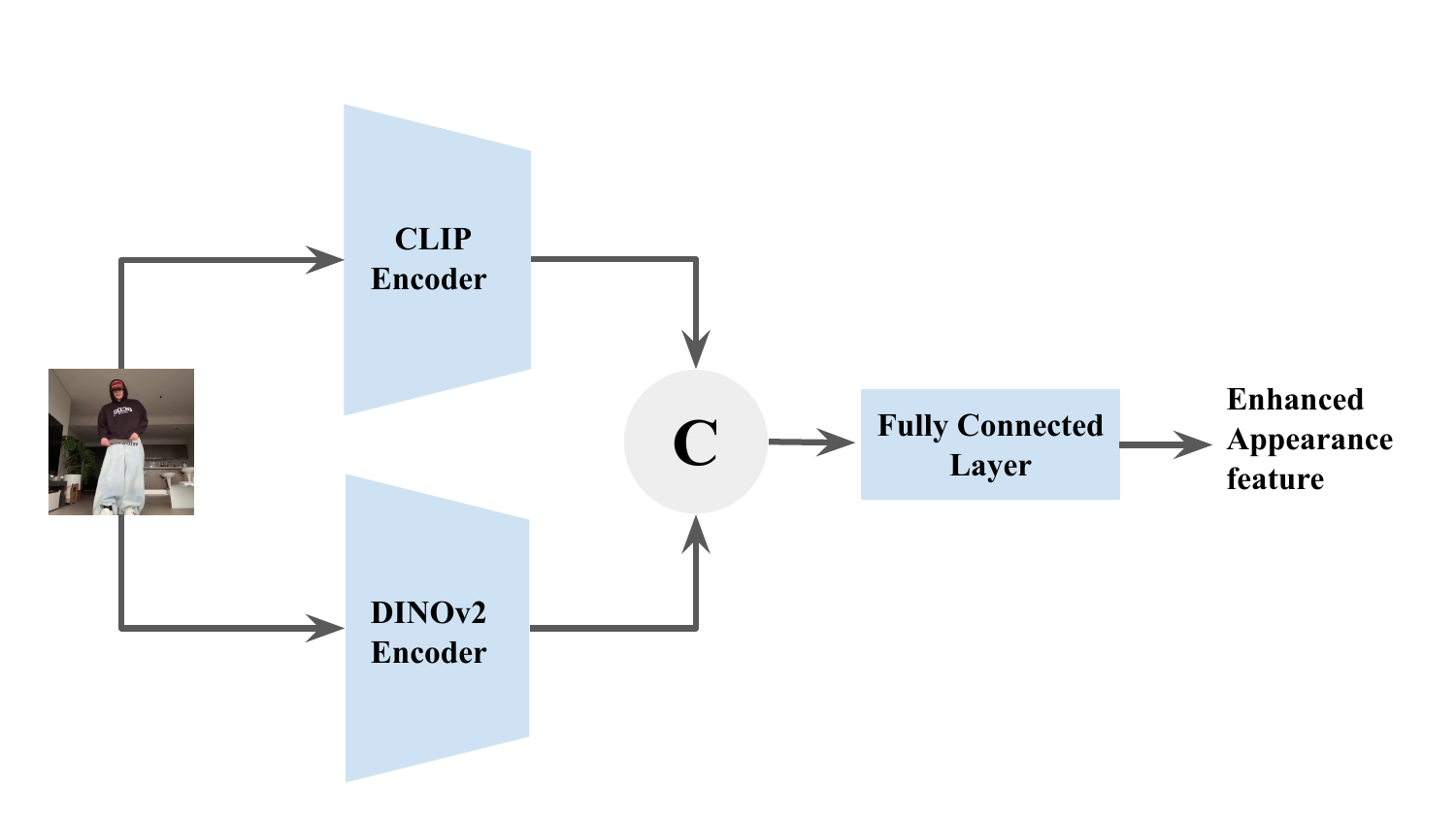}
    \caption{Illustration of our newly proposed Appearance Enhancement Module.} 
    \label{fig:AEM}
\end{figure}

Following the pipeline of text-guided conditional generation, image-to-video diffusion models in the early stage mainly use CLIP~\cite{radford2021learning} as their image condition encoder. However, as indicated in~\cite{karras2023dreampose}, CLIP is designed as a multi-modal vision and language model to connect texts and images, 
where the image encoder 
is designed to extract features that can better fit the text. 
Besides, CLIP 
focuses on 
abstracting the high-level 
information within the image. 
As a result, most of earlier studies on video generation only rely on the CLIP-based image condition, 
without paying attention to the appearance details in the target reference image. The reference image details are of vital importance when generating target video frames. 
Therefore, we can often observe 
the missing of details or the distortion in the dance videos generated by current works, especially in the facial area, limb distal segments of the dancers, and the background area around them. 


To solve the problem, we propose a novel module, Appearance Enhancement Module (AEM, $\mathbf{F}_{AEM}(\cdot)$), to 
better extract the details of the reference image 
so they can be applied to enhance the extracted appearance condition $c_{app}$. 
Specifically, as shown in Fig.~\ref{fig:AEM}, our AEM is composed of two main parts. Besides the commonly used CLIP encoder $\mathbf{E}_{clip}(\cdot)$, we also incorporate another ViT~\cite{han2021transformer}-based encoder, and a DINO-v2~\cite{oquab2023dinov2} ($\mathbf{E}_{dino}(\cdot)$), into our module. Different from CLIP, DINO-v2 is designed for the image segmentation task, which means it can handle the low-level information better and 
can be exploited to capture more details from the reference image $I_{tgt}$. In other words, these two components can provide information in two aspects: 
\begin{equation}
    \begin{aligned}
        c_{h} &= \mathbf{E}_{clip}(I_{tgt}), \\
        c_{l} &= \mathbf{E}_{dino}(I_{tgt}).~\label{eq:high&low}
    \end{aligned}
\end{equation}
By combining the extracted features from both encoders, we can ensure that both high-level semantic features $c_{h}$ and the low-level details $c_{l}$ 
from the target reference image can be simultaneously captured by our model. 
With the equal distribution of effort on the low-level details 
and the high-level abstract information, our AEM can provide an enhanced appearance condition $c_{app}$ 
through the effective combination of $c_{h}$ and $c_{l}$ using a fusion module $\mathbf{C}(\cdot)$, which is implemented with Fully-Connected Layers, 
and the total process within our AEM can be formulated as
\begin{equation}
    \begin{aligned}
        c_{app} &= \mathbf{F}_{AEM}(I_{tgt}) \\
                       &= \mathbf{C}(Concat[c_{h}, c_{l}]) \\
                       &= \mathbf{C}(Concat[\mathbf{E}_{clip}(I_{tgt}), \mathbf{E}_{dino}(I_{tgt})]).~\label{eq:aem}
    \end{aligned}
\end{equation}

\subsection{Pose Rendering Module (PRM)~\label{sec:3.4}}

\begin{figure}[t]
    \centering
    \includegraphics[width=\linewidth]{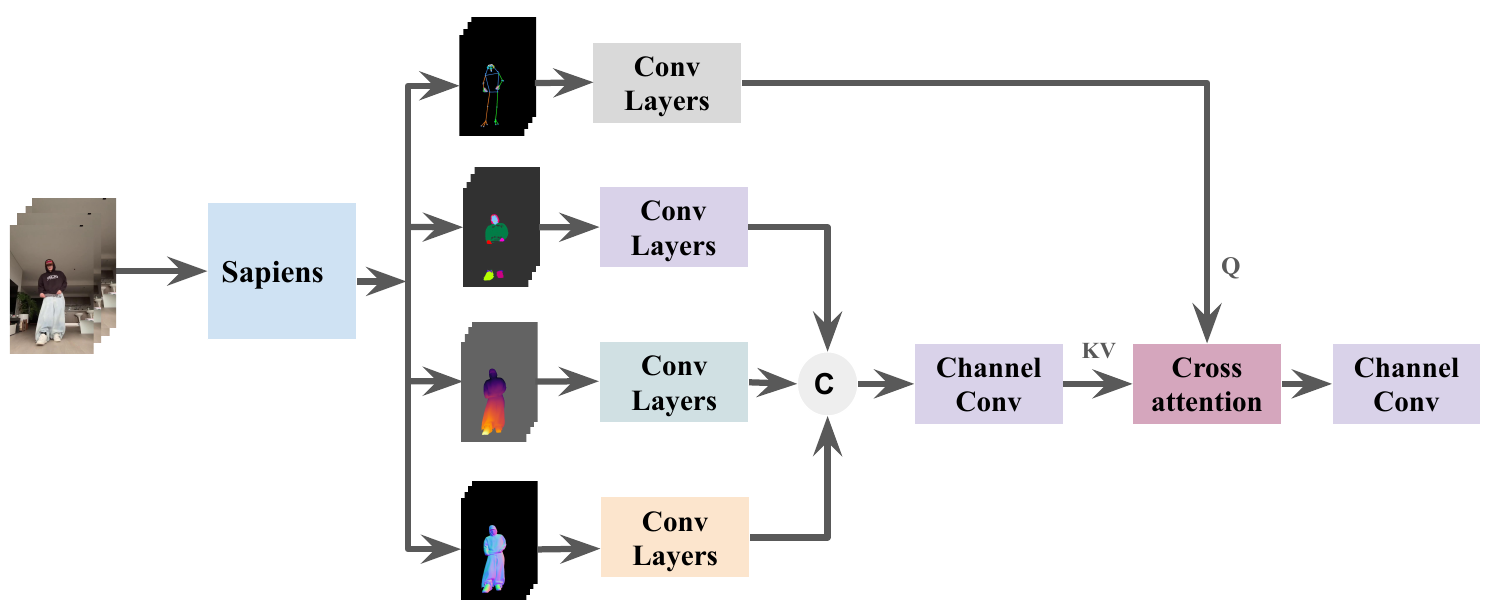}
    \caption{Illustration of our newly proposed Pose Rendering Module.}
    \label{fig:PRM}
\end{figure}


Another limitation observed in prior methods is the insufficient quality of pose guidance. Most existing approaches rely heavily on skeleton maps extracted from source dance videos to guide generation. However, inaccuracies in skeleton detection can significantly degrade generation performance. Furthermore, even dense skeletons \cite{yang2023effective}  fail to fully capture the richness of motion present in the source video, let alone the sparse skeletons commonly used by many existing models. As a result, the generated videos often lack precision and realism in pose representation and movement continuity. 

In order to capture the motion information in the source video $V_{src}$ to the greatest extent, we introduce a more effective Pose Rendering Module (PRM, $\mathbf{F}_{PRM}(\cdot)$), as shown in Fig.~\ref{fig:PRM}, to provide the motion condition $c_{pose}$ in Eq.~\eqref{eq:diffusion} to control the generation process. 
Specifically, we adopt \textbf{Sapiens} [16], a most recent model, as our pose extractor. 
Pre-trained on a large-scale dataset of human images, it has shown impressive capability on several human-centric tasks. For each frame $I_{src, i}$ in $V_{src}$, Sapiens model $\mathbf{E}_{sapiens}(\cdot)$ can provide pose information from different aspects, including a body-part segmentation map $P^{seg}_{src, i}$, a depth map $P^{dep}_{src, i}$, a normal map $P^{norm}_{src, i}$ as well as a skeleton map $P^{ske}_{src, i}$, 
and these components can complement each other to more effectively guide the video generation. In formula, the pose representation $P_{src, i} = \{P^{ske}_{src, i}, P^{seg}_{src, i}, P^{dep}_{src, i}, P^{norm}_{src, i}\}$ for the $i^{th}$ frame $I_{src, i}$ in the source video can be expressed as
\begin{equation}
    P_{src, i} = \mathbf{E}_{sapiens}(I_{src, i}).~\label{eq:sapiens}
\end{equation}

Different from~\cite{zhu2024champ}, where a 
single model is used to extract features from source inputs of all domains, we design separate convolutional blocks $\mathbf{Conv}_{ske}(\cdot)$, $\mathbf{Conv}_{seg}(\cdot)$, $\mathbf{Conv}_{dep}(\cdot)$, $\mathbf{Conv}_{norm}(\cdot)$ for different pose maps within the pose representation $P_{src, i}$. By doing so, we can guarantee that each encoder best fits the corresponding pose map. 
Incorporating the extracted features from augmented pose maps, we can obtain enriched pose guidance from different perspective, and the final motion condition $c_{pose} = \{c_{pos, 0}, c_{pos, 1}, ..., c_{pos, N}\}$ participating in the diffusion process in Eq.~\eqref{eq:diffusion} can be formulated as 
\begin{equation}
    \begin{aligned}
        c_{pos, i} &= \mathbf{H}_{2}(\mathbf{A}(c_{ske}, c_{aug})) \\
        &= \mathbf{H}_{2}(\mathbf{A}(c_{ske}, \mathbf{H}_{1}(Concat[c_{seg}, c_{dep}, c_{norm}]))), \\
        c_{ske} &= \mathbf{Conv}_{ske}(P^{ske}_{src, i}), \\
        c_{seg} &= \mathbf{Conv}_{seg}(P^{seg}_{src, i}), \\
        c_{dep} &= \mathbf{Conv}_{dep}(P^{dep}_{src, i}), \\
        c_{norm} &= \mathbf{Conv}_{norm}(P^{norm}_{src, i}),~\label{eq:c_motion}
    \end{aligned}
\end{equation}
where $\mathbf{A(\cdot)}$ is the cross-attention function, $\mathbf{H}_{1}(\cdot)$ and $\mathbf{H}_{2}(\cdot)$ are channel-wise convolutional block to further reduce the channel dimension of $c_{pos, i}$. 




\section{Experiments~\label{sec:4}}

\subsection{Experimental Settings~\label{sec:4.1}}


\begin{table*}[!htpb]
    \centering
    \caption{Overall performance of different models on the \textit{TikTok} testset. $\dagger$ represents the results measured by ourselves by rerunning the corresponding models. For each metric, the best result is in \textbf{bold} and the second best is \underline{underlined}.~\label{tab:overall_performance}}
    \resizebox{\textwidth}{!}{%
    \begin{tabular}{c c c c c c c c}
        \toprule
            &\multicolumn{5}{c}{Image-Level} &\multicolumn{2}{c}{Video-Level}  \\
            \cmidrule(lr){2-6} \cmidrule(lr){7-8}
            &FID ($\downarrow$) &SSIM ($\uparrow$) &LPIPS ($\downarrow$) &PSNR ($\uparrow$) &L1 ($\downarrow$) &FID-VID ($\downarrow$) &FVD ($\downarrow$) \\
        \midrule
            DISCO~\cite{wang2024disco} &28.31 &0.674 &0.285 &16.68 &3.69E-04 &55.17 &267.75 \\
            MagicPose~\cite{chang2023magicpose} &\underline{25.50} &0.752 &0.292 &29.53 &8.10E-05 &46.30 &/ \\
            VividPose~\cite{wang2024vividpose} &31.89 &0.758 &0.261 &\underline{29.83} &\underline{6.89E-05} &18.81 &152.97 \\
            AnimateAnyone~\cite{hu2024animate} &/ &0.718 &0.285 &29.56 &/ &/ &171.90 \\
            MimicMotion$^{\dagger}$~\cite{zhang2024mimicmotion} &33.65 &0.740 &0.296 &29.08 &9.07E-05 &19.31 &206.46 \\
        \midrule
            DANCER (w/o TikTok-3K) & 26.56 & \underline{0.802} & \underline{0.249} & 29.13 & 7.25E-05 & \underline{16.58} & \underline{140.64} \\
            DANCER (w/ TikTok-3K) & \textbf{25.27} & \textbf{0.803} & \textbf{0.247} & \textbf{30.03} & \textbf{6.48E-05} & \textbf{13.66} & \textbf{129.45} \\
        \bottomrule
    \end{tabular}
    }
\end{table*}

\subsubsection{Evaluation Metrics~\label{sec:4.1.2}}

To better evaluate the quality of generated dance videos, following outlines proposed in DisCo~\cite{wang2024disco}, we adopt metrics from two aspects: image generation quality and video generation quality. 
For the first part, we test frame-wise FID~\cite{heusel2017gans}, SSIM~\cite{wang2004image}, LPIPS~\cite{zhang2018unreasonable}, PSNR~\cite{hore2010image} and L1; while for the second part, we report FID-VID~\cite{balaji2019conditional} and FVD~\cite{unterthiner2018towards} for every consecutive $16$ frames. 

\subsubsection{Implementation Details~\label{sec:4.1.3}}
We adopt the pre-trained stable video diffusion image-to-video model. During training, we train the weights of U-Net~\cite{ronneberger2015u} and our augmented pose module, keeping other parts 
frozen. All experiments are conducted on 8 NVIDIA H100 with batch-size 1 and learning rate 1e-5. Each batch includes 1 reference frame, 8 randomly sampled consecutive ground-truth frames and corresponding extracted pose images.


\subsection{Overall Performance~\label{sec:4.2}}

In order to quantitatively evaluate our model, we conduct experiments on \textit{TikTok}~\cite{jafarian2021learning} dataset and compare the performance with some other baselines, including DISCO~\cite{wang2024disco}, MagicPose~\cite{chang2023magicpose}, VividPose~\cite{wang2024vividpose}, AnimateAnyone~\cite{hu2024animate} and MimicMotion~\cite{zhang2024mimicmotion}.
During experiments, we conduct the same preprocessing as done in~\cite{wang2024disco} to data for fair comparison. The overall results are provided in Table.~\ref{tab:overall_performance}. 
We can see that our proposed model is superior to the state-of-the-art methods on the single-person dance synthesis task with obviously better SSIM, LPIPS, FID-VID and FVD. Specifically, our SSIM surpasses Vividpose by 5.8\%, LPIPS by 4.6\%, FID-VID by 12\% and FVD by 8\%. The improvement is reflected from two aspects. First, according to the image-level metrics, our proposed model can generate high-quality video frames, which is attributed to the attention from our model to the details of reference images. Second, the videos generated by our model have a better continuity, which is demonstrated through its improvement of video-level measurement.


    \begin{table*}[!htpb]
        \centering
        \caption{Performance comparison among our model with different appearance encoders on \textit{TikTok} dataset. For each metric, the best result is in \textbf{bold}.}
        \label{tab:appearance}
        \resizebox{\textwidth}{!}{%
        \begin{tabular}{c c c c c c c c c c}
            \toprule
                \multicolumn{2}{c}{AEM} &\multicolumn{5}{c}{Image-Level} &\multicolumn{2}{c}{Video-Level}  \\
                \cmidrule(lr){1-2} \cmidrule(lr){3-7} \cmidrule(lr){8-9}
                CLIP &DINO-v2 &FID ($\downarrow$) &SSIM ($\uparrow$) &LPIPS ($\downarrow$) &PSNR ($\uparrow$) &L1 ($\downarrow$) &FID-VID ($\downarrow$) &FVD ($\downarrow$) \\
            \midrule 
                \cmark &\xmark & 32.49 & 0.7693 & 0.2874 & \textbf{29.37} & 8.27E-5 & 22.34 & 167.06 \\
                \cmark &\cmark & \textbf{31.53} & \textbf{0.7782} & \textbf{0.2604} & 29.12 & \textbf{7.91E-5} & \textbf{21.95} & \textbf{159.33} \\
            \bottomrule
        \end{tabular}
        }
    \end{table*}

    \begin{table*}[!htpb]
        \centering
        \caption{Performance comparison between augmented pose guidance and traditional skeleton map on \textit{TikTok} dataset. For each metric, the best result is in \textbf{bold}.}
        \label{tab:motion}
        \resizebox{\textwidth}{!}{%
        \begin{tabular}{c c c c c c c c c c}
            \toprule
                \multicolumn{2}{c}{PRM} &\multicolumn{5}{c}{Image-Level} &\multicolumn{2}{c}{Video-Level}  \\
                \cmidrule(lr){1-2} \cmidrule(lr){3-7} \cmidrule(lr){8-9}
                Skeleton &Sapiens &FID ($\downarrow$) &SSIM ($\uparrow$) &LPIPS ($\downarrow$) &PSNR ($\uparrow$) &L1 ($\downarrow$) &FID-VID ($\downarrow$) &FVD ($\downarrow$) \\
            \midrule
                \cmark &\xmark & 32.49 & 0.7693 & 0.2874 & \textbf{29.37} & 8.27E-5 & 22.34 & 167.06  \\
                \cmark &\cmark & \textbf{29.97} & \textbf{0.7728} & \textbf{0.2639} & 29.31 & \textbf{7.34E-5} & \textbf{19.03} & \textbf{155.02} \\
            \bottomrule
        \end{tabular}
        }
    \end{table*}

\subsection{Ablation Study~\label{sec:4.3}}

To further demonstrate the effectiveness of our design, we separately study each component through extensive experiments, including the \textit{Appearance Enhancement Module}, the \textit{Pose Rendering Module} and the effect of additional training data brought by our novel dataset.

\begin{figure*}[!htb]
    \centering
    \begin{subfigure}{.13\linewidth}
        \centering
        \caption*{\makebox[7em][c]{Reference}}
        \includegraphics[width=\linewidth]{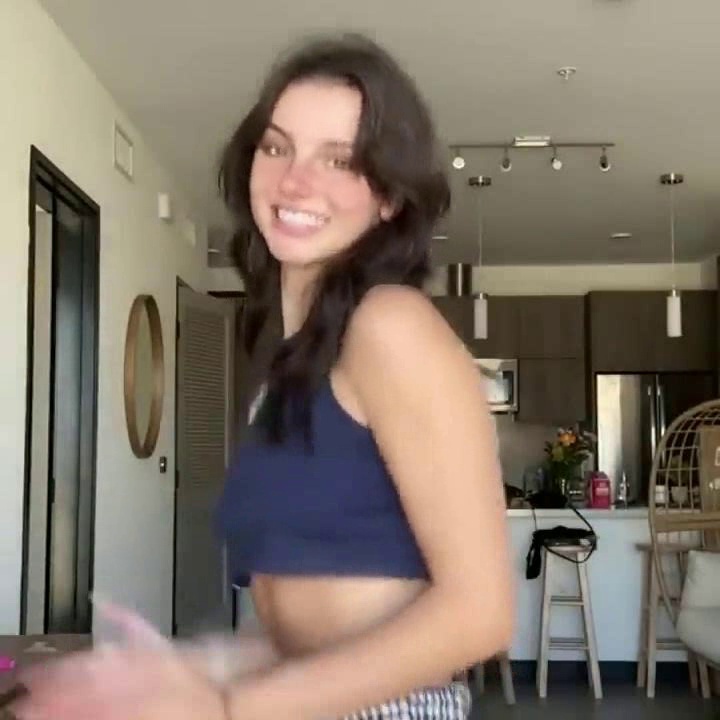}
    \end{subfigure}
    \hfill
    \begin{subfigure}{.13\linewidth}
        \centering
        \caption*{\makebox[7em][c]{DWPose}}
        \includegraphics[width=\linewidth]{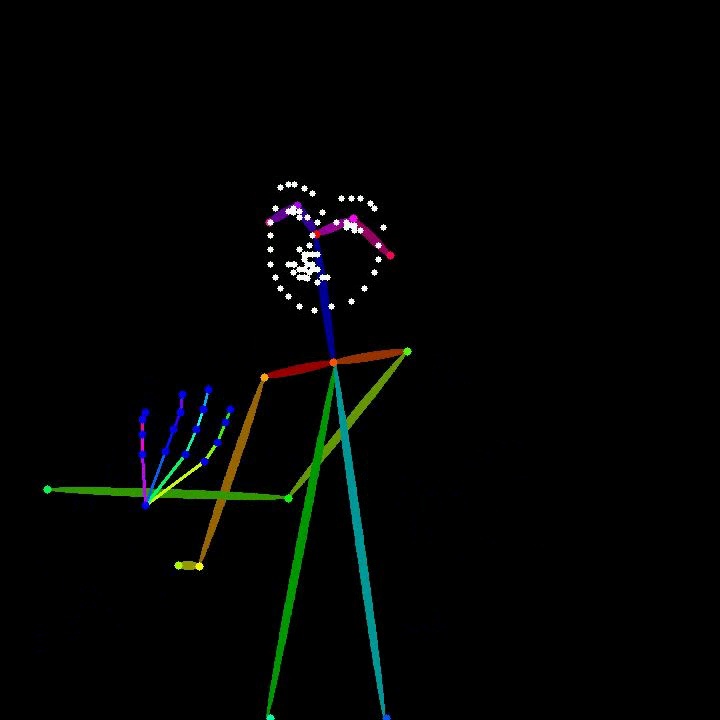}
    \end{subfigure}
    \hfill
    \begin{subfigure}{.13\linewidth}
        \centering
        \caption*{\makebox[7em][c]{AnimateAnyone}}
        \includegraphics[width=\linewidth]{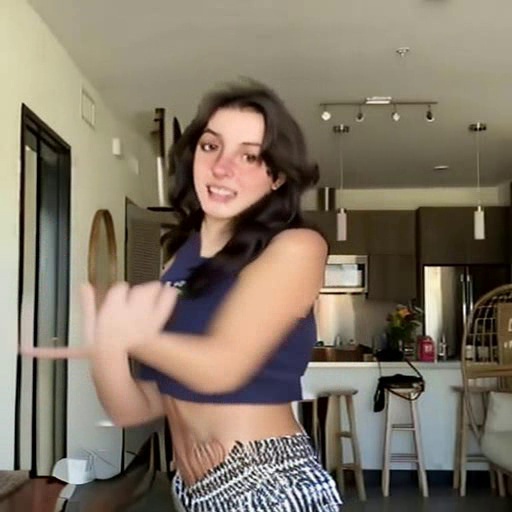}
    \end{subfigure}
    \hfill
    \begin{subfigure}{.13\linewidth}
        \centering
        \caption*{\makebox[7em][c]{MagicPose}}
        \includegraphics[width=\linewidth]{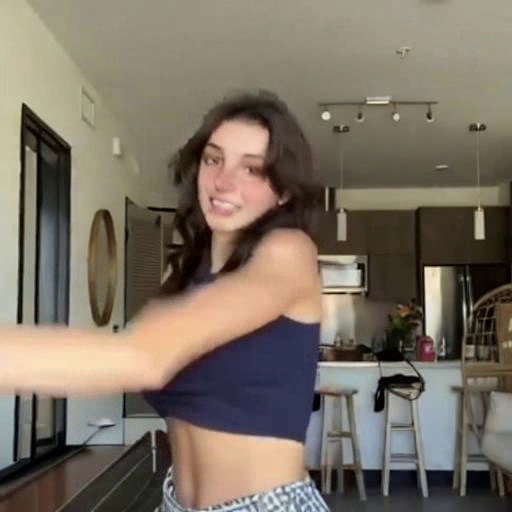}
    \end{subfigure}
    \hfill
    \begin{subfigure}{.13\linewidth}
        \centering
        \caption*{\makebox[7em][c]{MimicMotion}}
        \includegraphics[width=\linewidth]{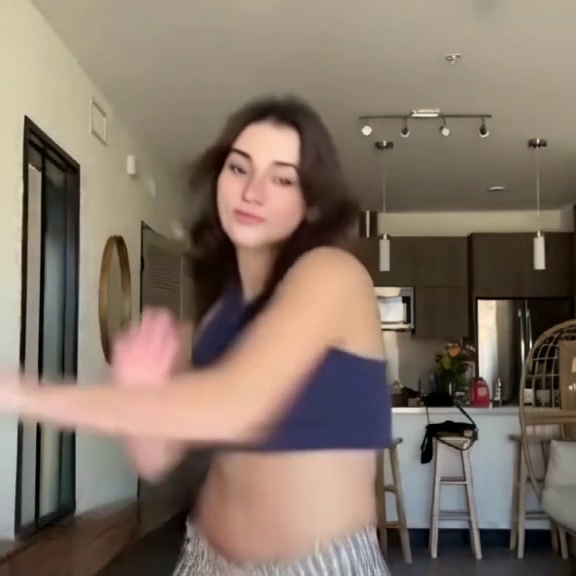}
    \end{subfigure}
    \hfill
    \begin{subfigure}{.13\linewidth}
        \centering
        \caption*{\makebox[7em][c]{SapiensPose}}
        \includegraphics[width=\linewidth]{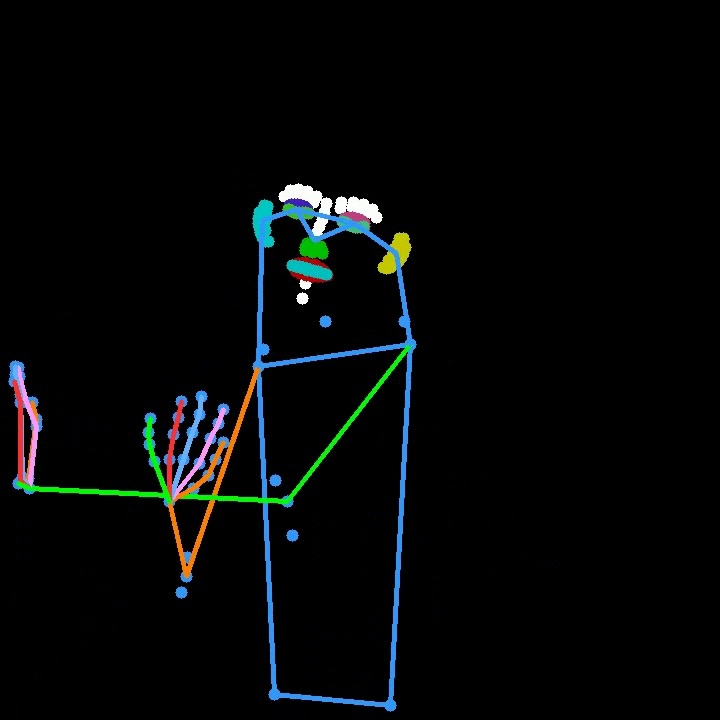}
    \end{subfigure}
    \hfill
    \begin{subfigure}{.13\linewidth}
        \centering
        \caption*{\textbf{Ours}}
        \includegraphics[width=\linewidth]{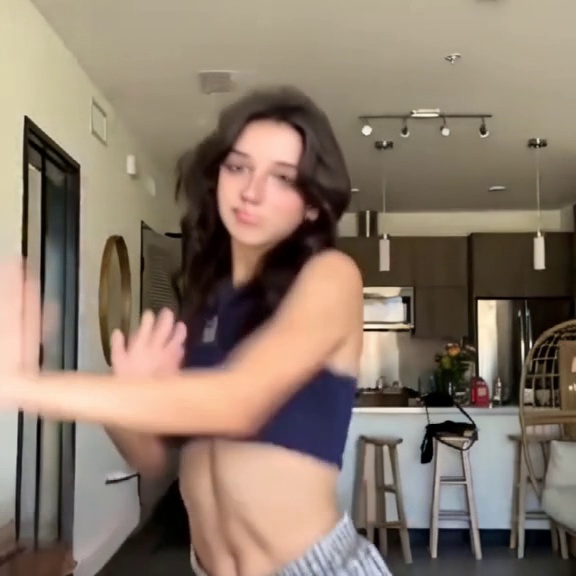}
    \end{subfigure}
    \begin{subfigure}{.13\linewidth}
        \centering
        \includegraphics[width=\linewidth]{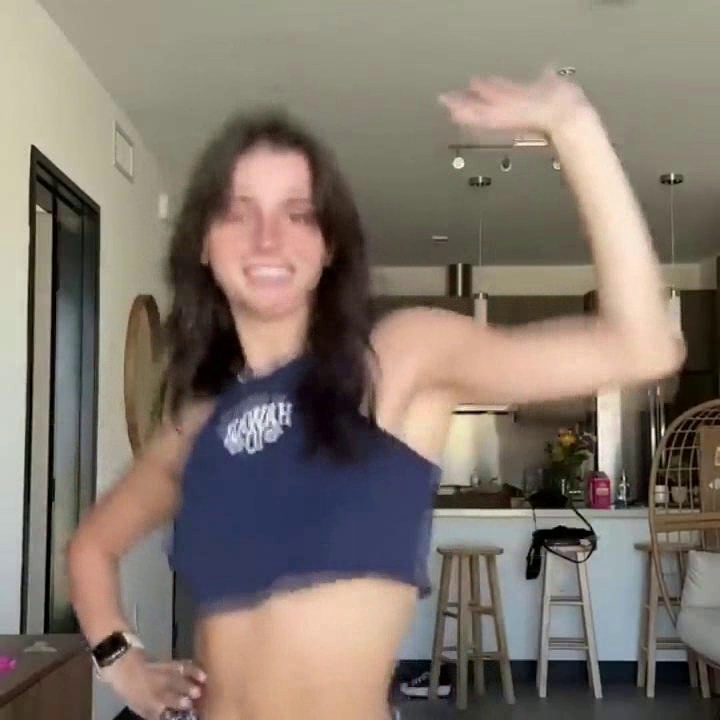}
    \end{subfigure}
    \hfill
    \begin{subfigure}{.13\linewidth}
        \centering
        \includegraphics[width=\linewidth]{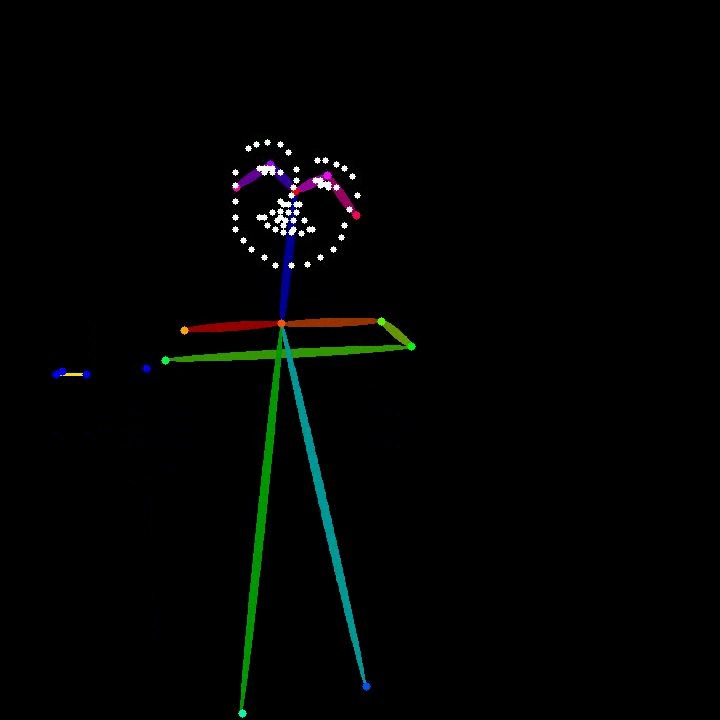}
    \end{subfigure}
    \hfill
    \begin{subfigure}{.13\linewidth}
        \centering
        \includegraphics[width=\linewidth]{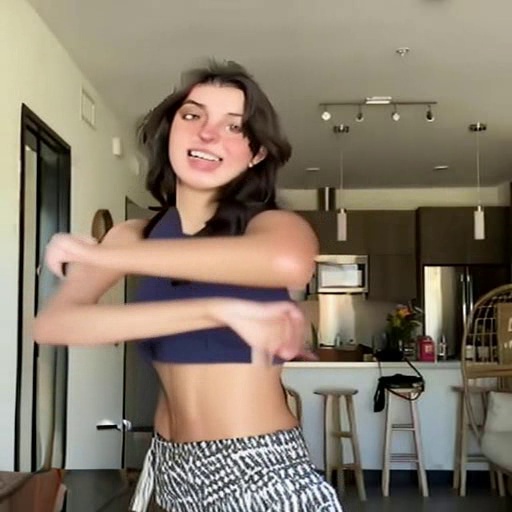}
    \end{subfigure}
    \hfill
    \begin{subfigure}{.13\linewidth}
        \centering
        \includegraphics[width=\linewidth]{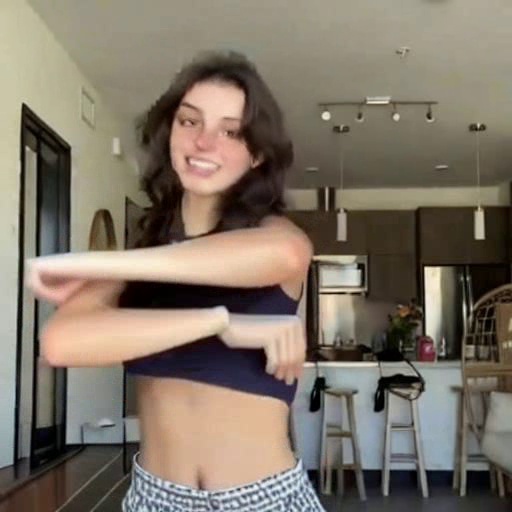}
    \end{subfigure}
    \hfill
    \begin{subfigure}{.13\linewidth}
        \centering
        \includegraphics[width=\linewidth]{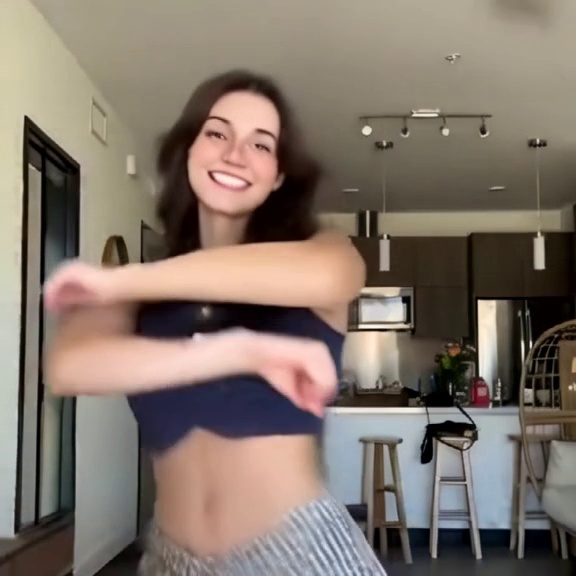}
    \end{subfigure}
    \hfill
    \begin{subfigure}{.13\linewidth}
        \centering
        \includegraphics[width=\linewidth]{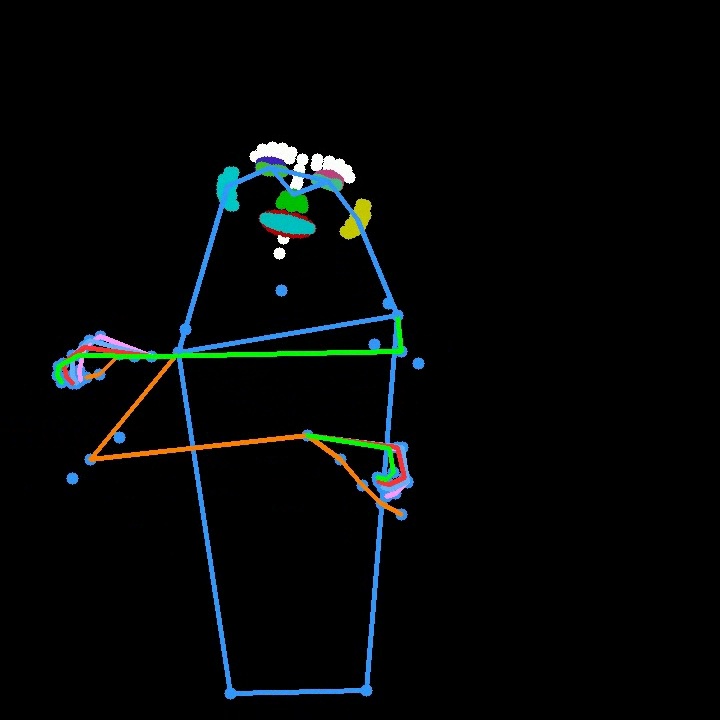}
    \end{subfigure}
    \hfill
    \begin{subfigure}{.13\linewidth}
        \centering
        \includegraphics[width=\linewidth]{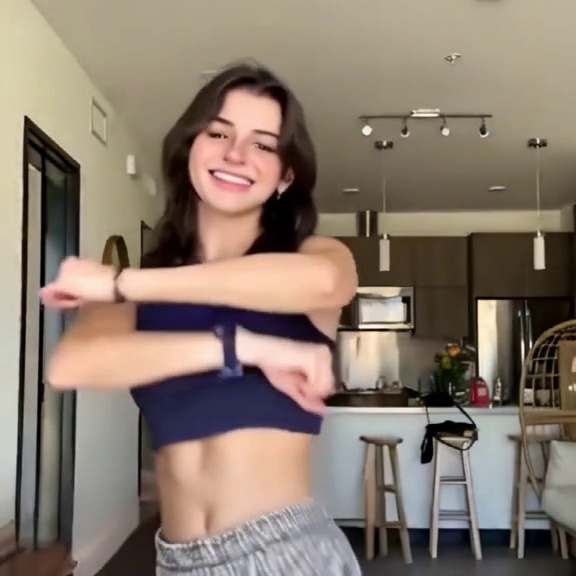}
    \end{subfigure}
    \begin{subfigure}{.13\linewidth}
        \centering
        \includegraphics[width=\linewidth]{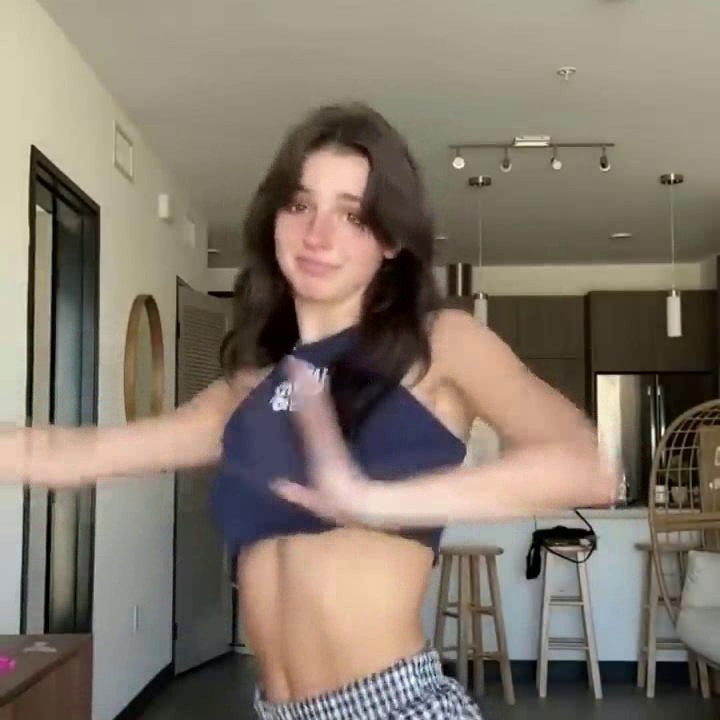}
    \end{subfigure}
    \hfill
    \begin{subfigure}{.13\linewidth}
        \centering
        \includegraphics[width=\linewidth]{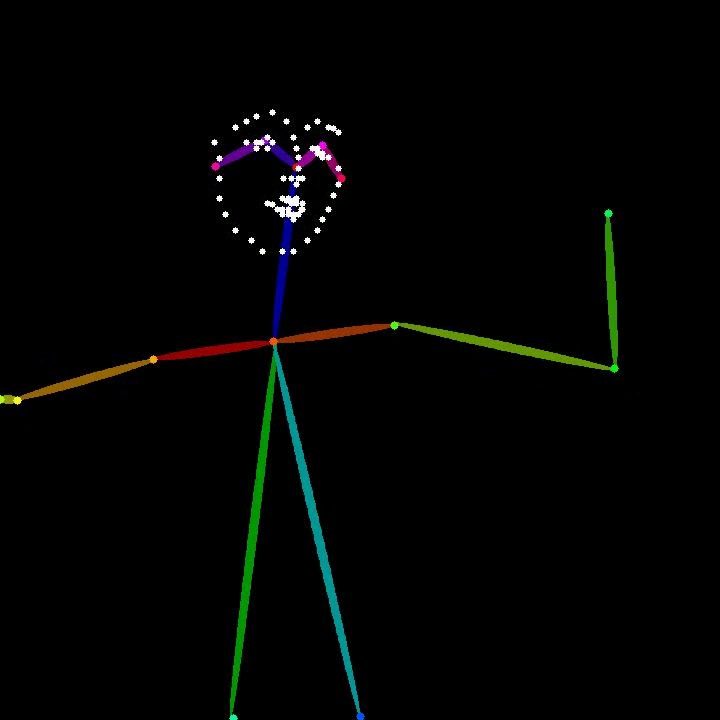}
    \end{subfigure}
    \hfill
    \begin{subfigure}{.13\linewidth}
        \centering
        \includegraphics[width=\linewidth]{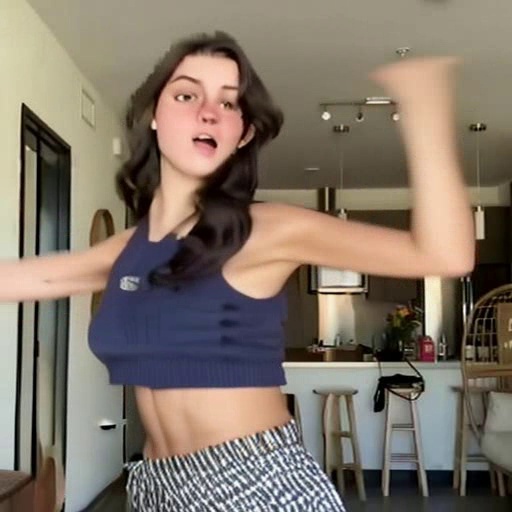}
    \end{subfigure}
    \hfill
    \begin{subfigure}{.13\linewidth}
        \centering
        \includegraphics[width=\linewidth]{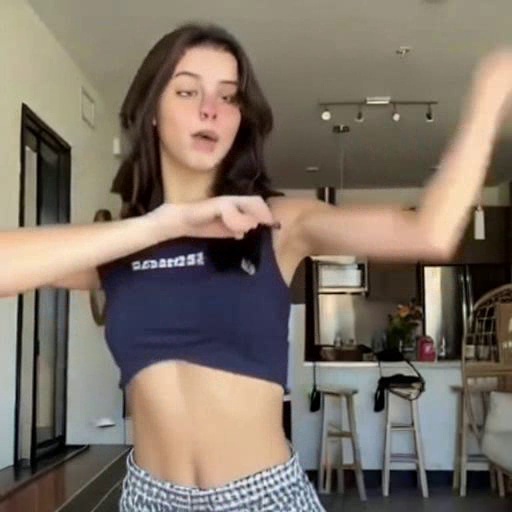}
    \end{subfigure}
    \hfill
    \begin{subfigure}{.13\linewidth}
        \centering
        \includegraphics[width=\linewidth]{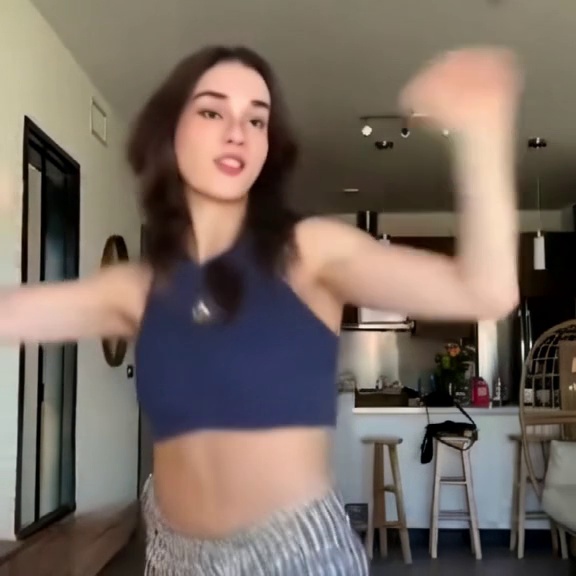}
    \end{subfigure}
    \hfill
    \begin{subfigure}{.13\linewidth}
        \centering
        \includegraphics[width=\linewidth]{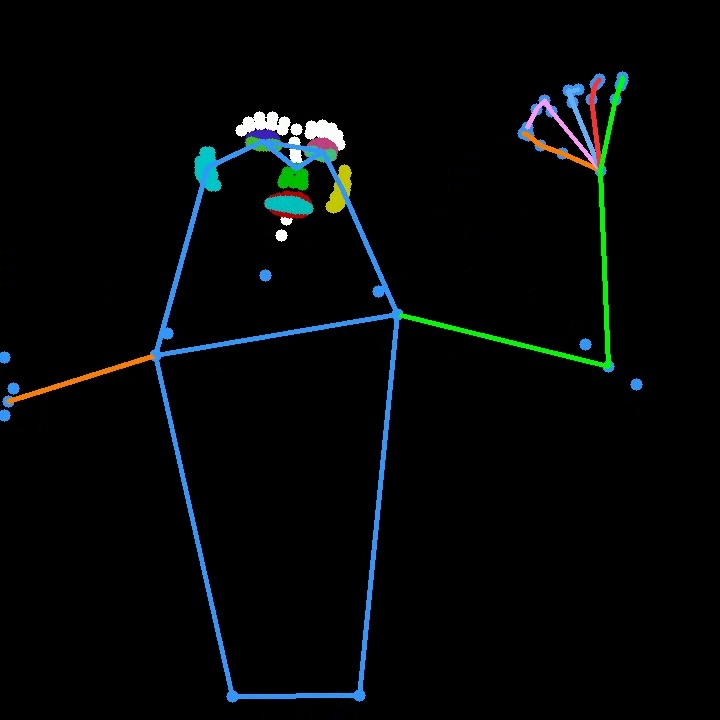}
    \end{subfigure}
    \hfill
    \begin{subfigure}{.13\linewidth}
        \centering
        \includegraphics[width=\linewidth]{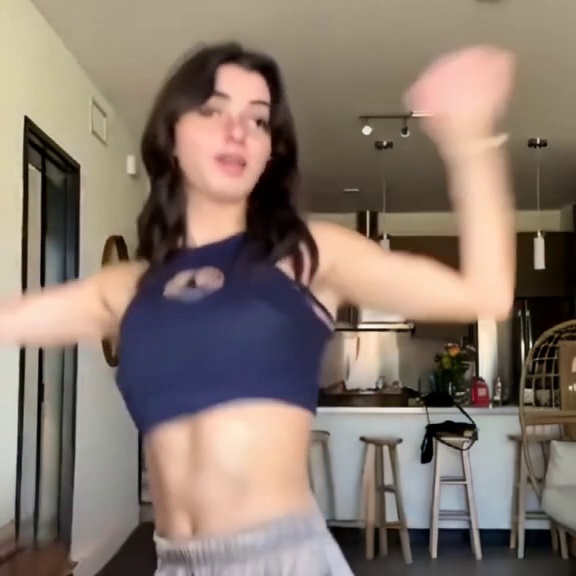}
    \end{subfigure}
    \begin{subfigure}{.13\linewidth}
        \centering
        \includegraphics[width=\linewidth]{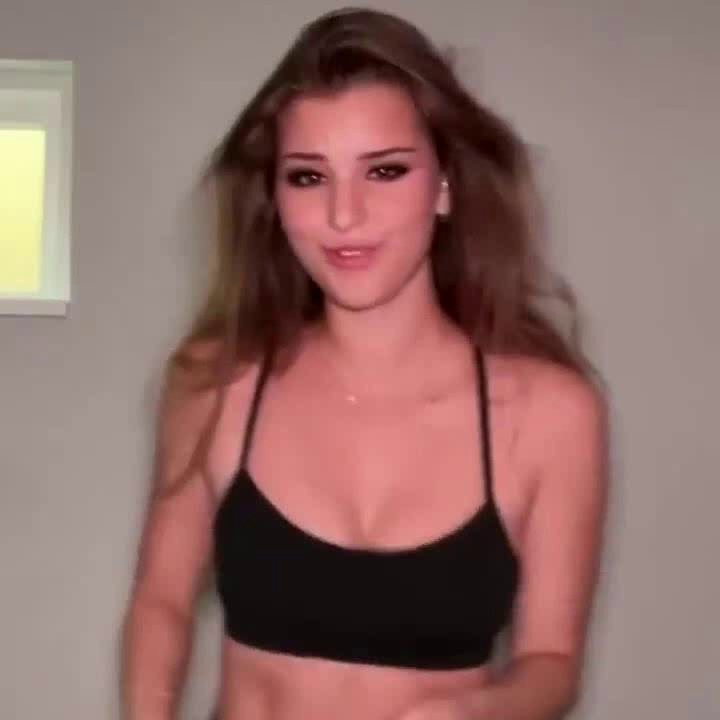}
    \end{subfigure}
    \hfill
    \begin{subfigure}{.13\linewidth}
        \centering
        \includegraphics[width=\linewidth]{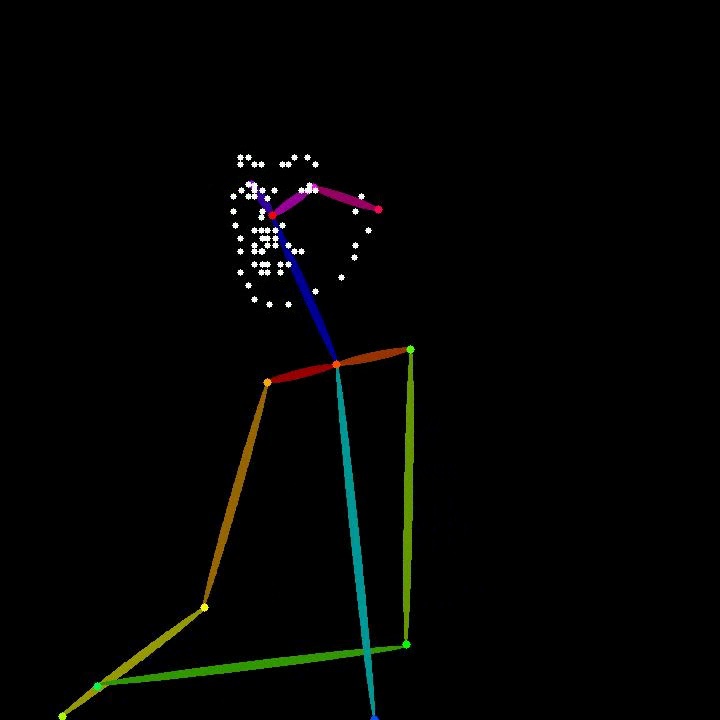}
    \end{subfigure}
    \hfill
    \begin{subfigure}{.13\linewidth}
        \centering
        \includegraphics[width=\linewidth]{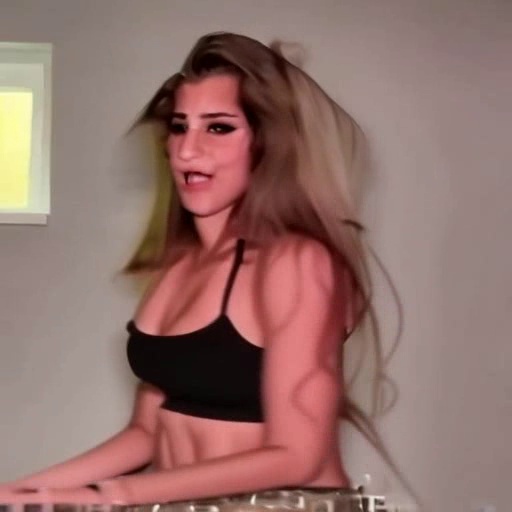}
    \end{subfigure}
    \hfill
    \begin{subfigure}{.13\linewidth}
        \centering
        \includegraphics[width=\linewidth]{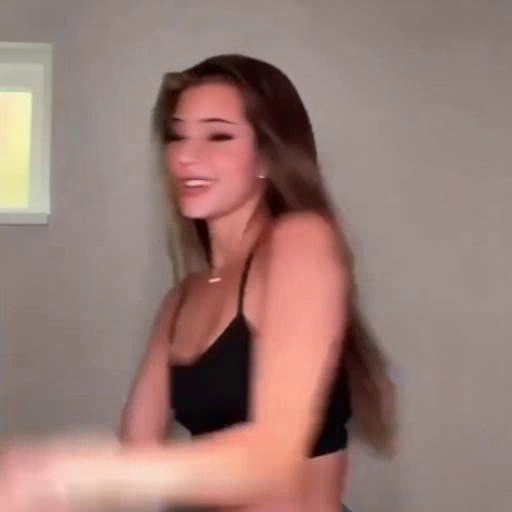}
    \end{subfigure}
    \hfill
    \begin{subfigure}{.13\linewidth}
        \centering
        \includegraphics[width=\linewidth]{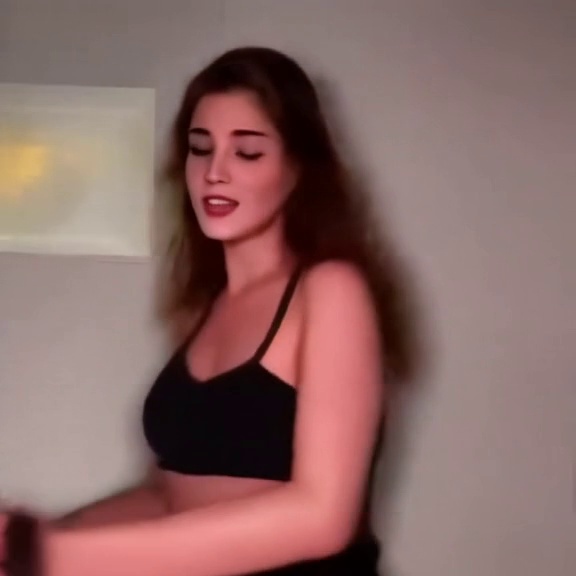}
    \end{subfigure}
    \hfill
    \begin{subfigure}{.13\linewidth}
        \centering
        \includegraphics[width=\linewidth]{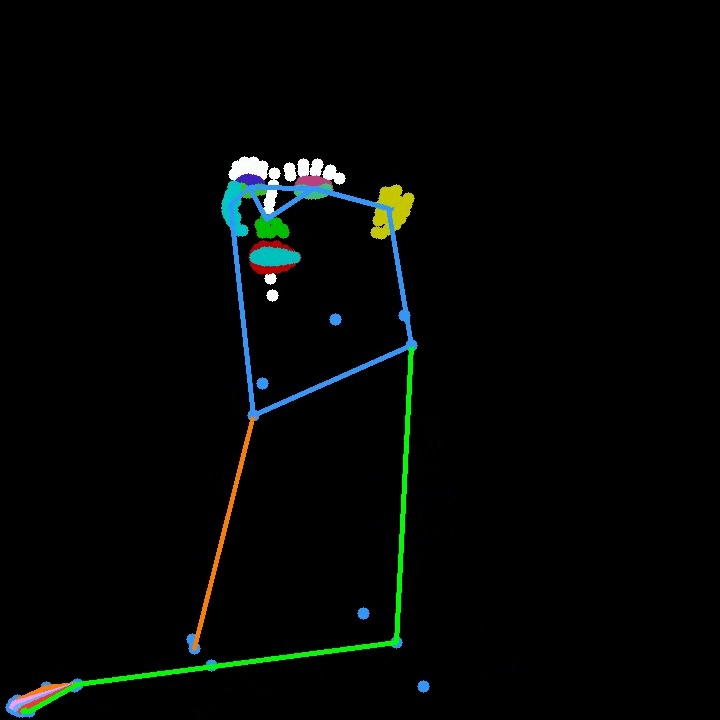}
    \end{subfigure}
    \hfill
    \begin{subfigure}{.13\linewidth}
        \centering
        \includegraphics[width=\linewidth]{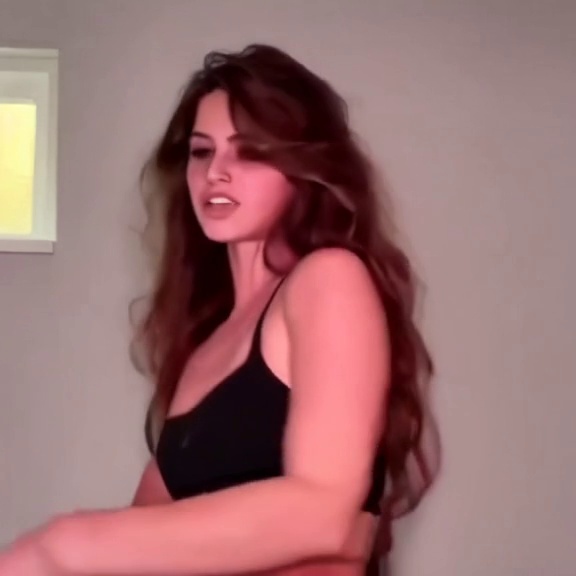}
    \end{subfigure}
    \begin{subfigure}{.13\linewidth}
        \centering
        \includegraphics[width=\linewidth]{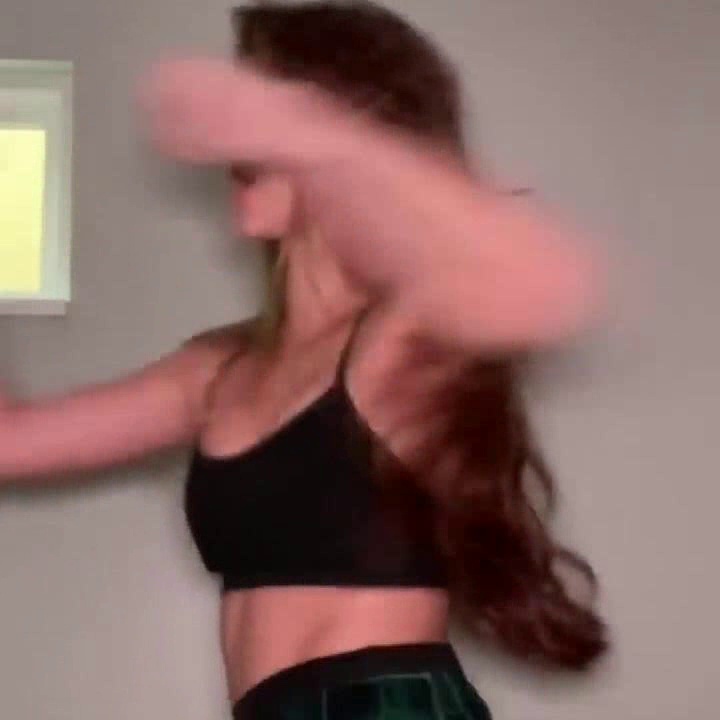}
    \end{subfigure}
    \hfill
    \begin{subfigure}{.13\linewidth}
        \centering
        \includegraphics[width=\linewidth]{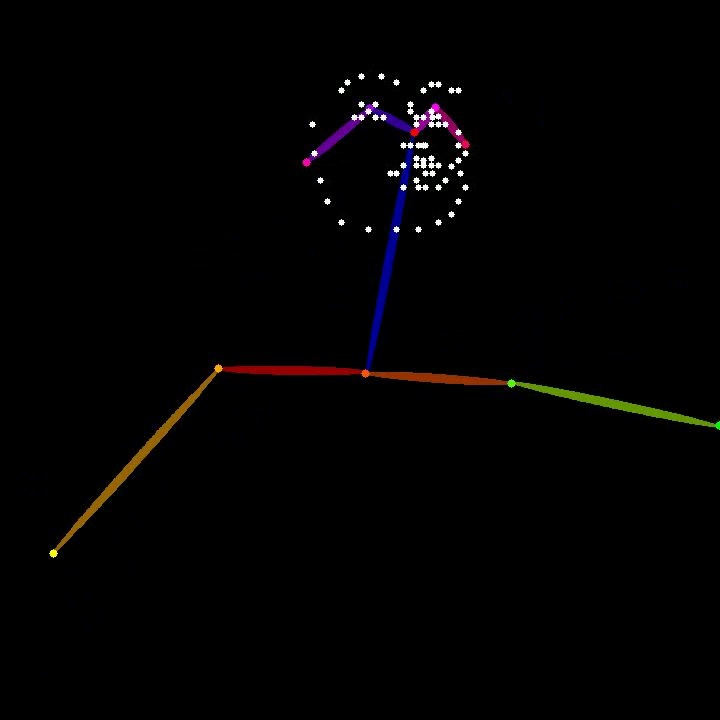}
    \end{subfigure}
    \hfill
    \begin{subfigure}{.13\linewidth}
        \centering
        \includegraphics[width=\linewidth]{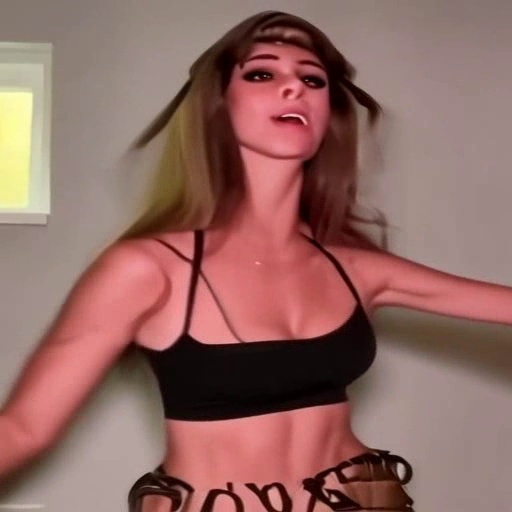}
    \end{subfigure}
    \hfill
    \begin{subfigure}{.13\linewidth}
        \centering
        \includegraphics[width=\linewidth]{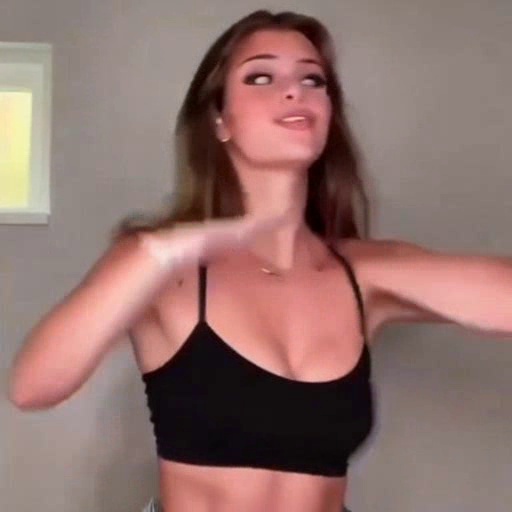}
    \end{subfigure}
    \hfill
    \begin{subfigure}{.13\linewidth}
        \centering
        \includegraphics[width=\linewidth]{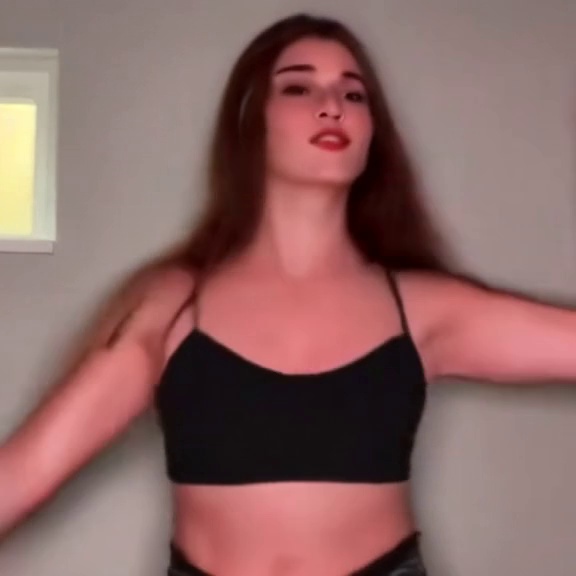}
    \end{subfigure}
    \hfill
    \begin{subfigure}{.13\linewidth}
        \centering
        \includegraphics[width=\linewidth]{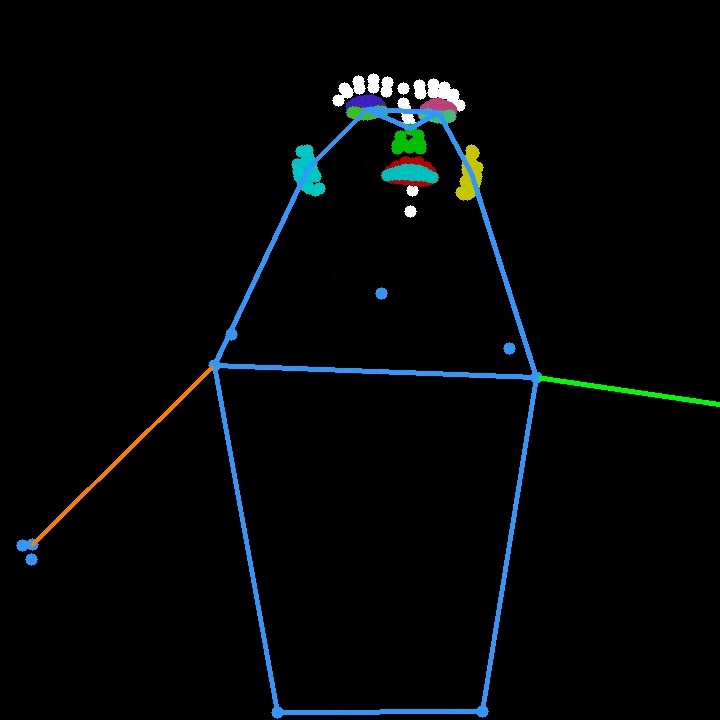}
    \end{subfigure}
    \hfill
    \begin{subfigure}{.13\linewidth}
        \centering
        \includegraphics[width=\linewidth]{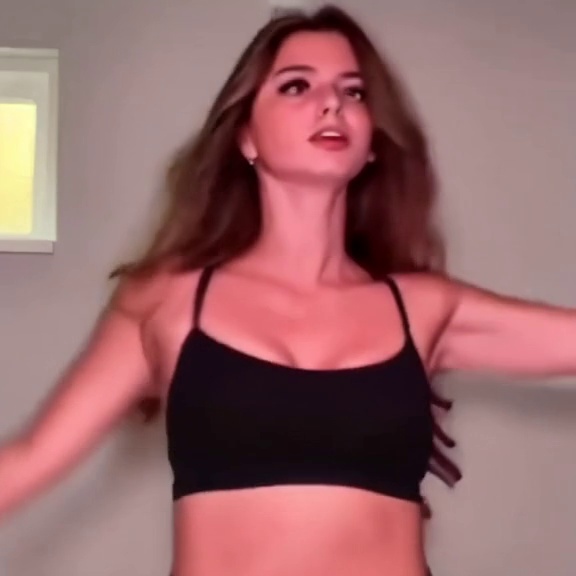}
    \end{subfigure}
    \begin{subfigure}{.13\linewidth}
        \centering
        \includegraphics[width=\linewidth]{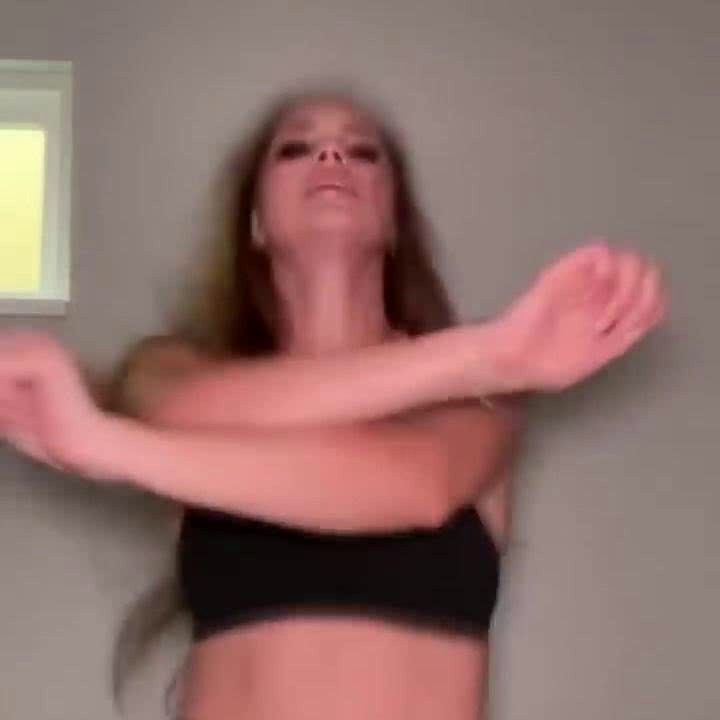}
    \end{subfigure}
    \hfill
    \begin{subfigure}{.13\linewidth}
        \centering
        \includegraphics[width=\linewidth]{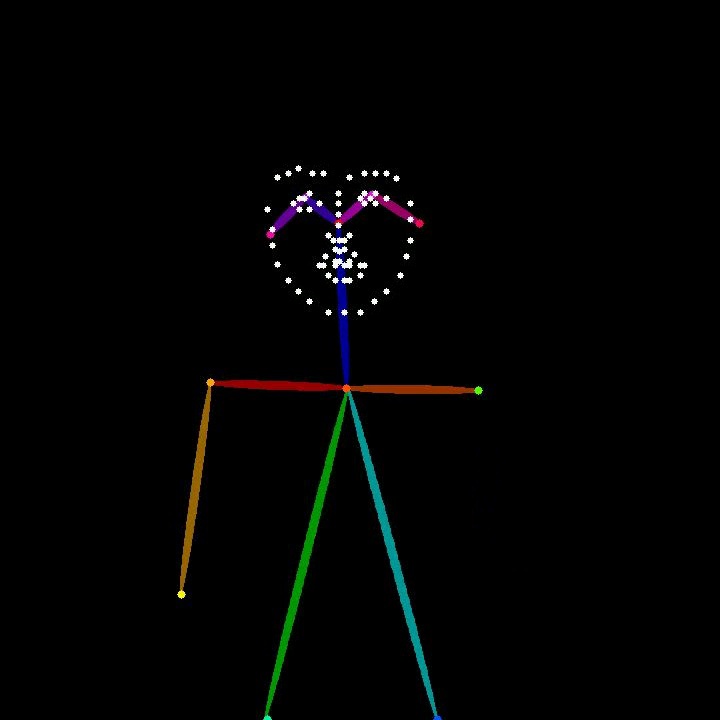}
    \end{subfigure}
    \hfill
    \begin{subfigure}{.13\linewidth}
        \centering
        \includegraphics[width=\linewidth]{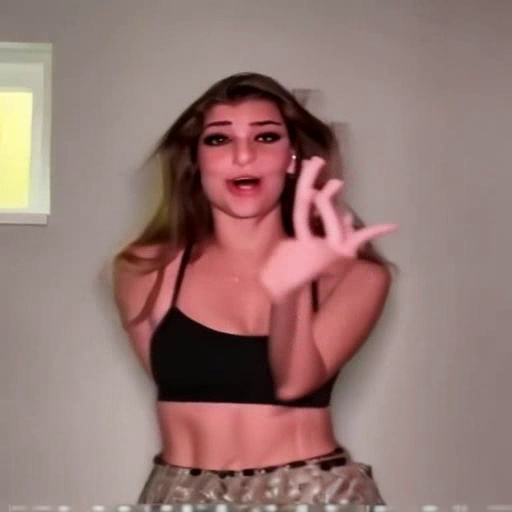}
    \end{subfigure}
    \hfill
    \begin{subfigure}{.13\linewidth}
        \centering
        \includegraphics[width=\linewidth]{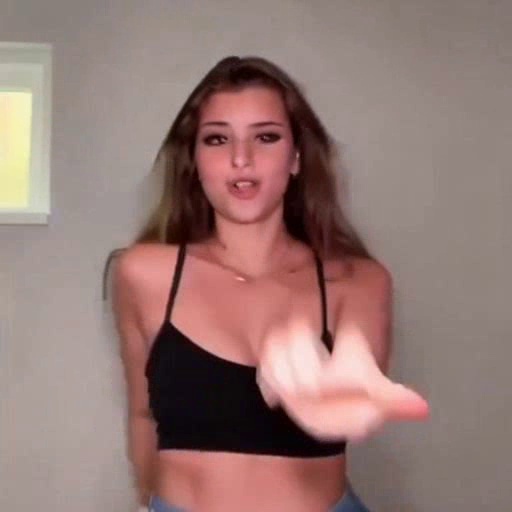}
    \end{subfigure}
    \hfill
    \begin{subfigure}{.13\linewidth}
        \centering
        \includegraphics[width=\linewidth]{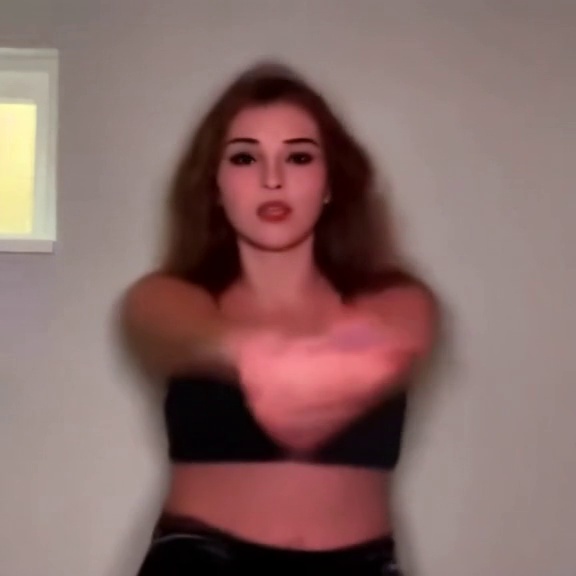}
    \end{subfigure}
    \hfill
    \begin{subfigure}{.13\linewidth}
        \centering
        \includegraphics[width=\linewidth]{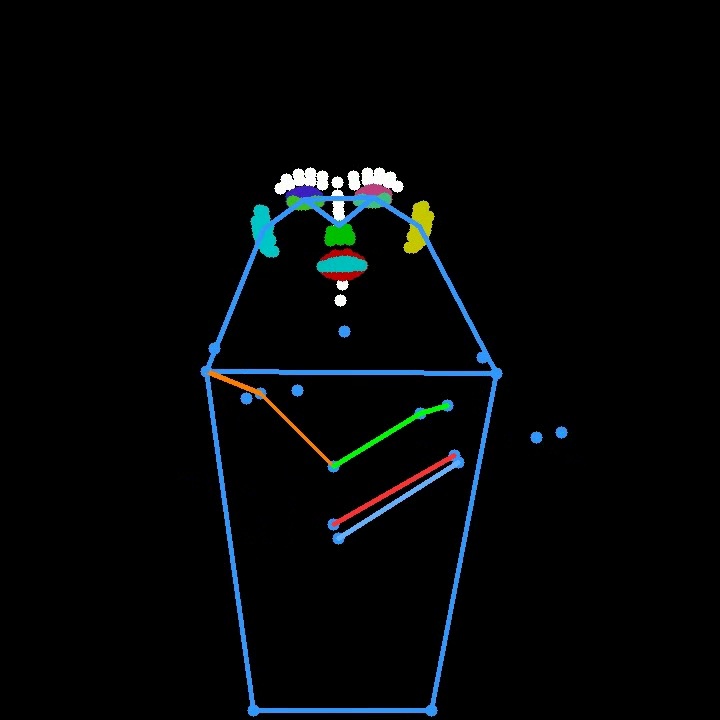}
    \end{subfigure}
    \hfill
    \begin{subfigure}{.13\linewidth}
        \centering
        \includegraphics[width=\linewidth]{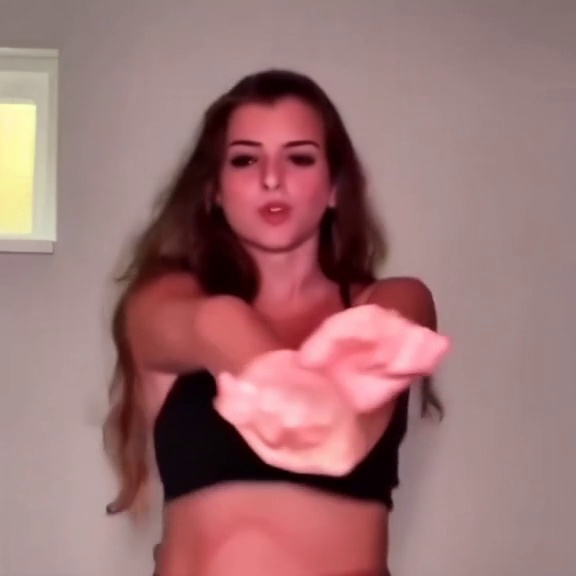}
    \end{subfigure}
    \caption{Illustration of the comparison among generated video samples using different methods.}
    \label{fig:qualititative}
\end{figure*}

\begin{itemize}
    \item \textbf{Appearance Enhancement Module:} In this part, we compare the performance of our framework using the newly designed appearance enhancement module (AEM) with the same framework only containing CLIP encoder. As shown in Table.~\ref{tab:appearance}, the effectiveness of our novel module has been proved through its better performance compared to others, which confirmed the importance of feature extraction in different levels and the extra attention on details. 
    A larger improvement is observed in the evaluation of the overall video performance, while the continuity is an important factor to consider. This is as expected, as a complete video involves more details and requires a method that is able to comprehensively consider all the information.   
    \item \textbf{Pose Rendering Module:} Another design that we want to evaluate is the Pose Rendering Module (PRM) in our model. As 
    discussed in Sec.~\ref{sec:3.4}, compared to only using skeleton maps, augmented pose guidance can provide extra information for reference. 
    According to Table.~\ref{tab:motion}, the improvement brought by the augmented pose guidance in our pose rendering module is obvious, especially in FID metric.

    \item \textbf{Additional Training Data:} To further demonstrate the impact of our newly proposed large-scale dataset, we also conduct extra experiments to compare the model performances with and without using our \textit{TikTok-3K} for further training. From the Table.~\ref{tab:overall_performance}, we can see that the continuing training on the new dataset obviously empowers our model, and the performance improvement is larger for the video sequence generation as it requires more data in the training. Since our \textit{TikTok-3K} contains much larger number of data samples, much longer video duration as well as much more diversity on dance contents, it can greatly improve the performance of machine learning algorithms that rely on data. The big dataset we create not only helps improve the performance of our model, but will also provide benefits to future methods exploring video related fields.

    
\end{itemize}

\subsection{Qualitative Study~\label{sec:4.4}}


To illustrate our results more effectively, we provide generated samples for comparison. As shown in Fig.~\ref{fig:qualititative}, the dance videos produced using our method, DANCER, exhibit superior visual quality compared to previous works, featuring clearer details. Specifically, the pose images obtained from Sapiens capture clearer body shapes, finer details, and more accurate orientation than those derived from DWPose, as evident in the first and last rows of Fig.~\ref{fig:qualititative}. This enhanced body rendering improves coherence with the reference object, significantly enhancing fidelity. Moreover, fine-grained details such as abdominal muscles, facial expressions, and hair shapes are well preserved, rather than being overly smoothed. These improvements are attributable to our intrinsic encoder design, which effectively synthesizes low-level structural features with high-level semantic features. Consequently, our results align more naturally with the reference, offering a more visually consistent outcome compared to prior approaches.




\section{Conclusion~\label{sec:5}}

In this paper, we propose a novel framework for human dance synthesis based on 
the novel generation of conditions to guide the 
operation of video diffusion models.  
To better leverage pose references, beyond traditional skeleton maps, we design a pose rendering module 
that incorporates diverse motion cues, including segmentation maps, depth maps, and normal maps.
As another important component of our proposed framework, we design a new appearance enhancement module to 
extract 
finer visual details 
from the reference image, 
such as facial features, limb distal segments and the background areas around the dancer. The effectiveness of our design has been proved on real-world datasets through extensive experiments. 
To further support model training and evaluation, 
we compile and plan to release a large-scale dance video dataset, which we believe will benefit the broader research community.
Although we achieve some promising results, we also see that potential challenges 
remain in this field. 
In the future, we will extend our current work to more complex scenarios like multi-person dance or general human activity generation.

\bibliographystyle{ACM-Reference-Format}
\bibliography{acmmm25}

\end{document}